\documentclass[lettersize,journal]{IEEEtran}
\usepackage[colorlinks=true, allcolors=blue]{hyperref}
\usepackage{amsmath,amsfonts}
\usepackage{amssymb}
\usepackage{algorithmic}
\usepackage{algorithm}
\usepackage{array}
\usepackage{textcomp}
\usepackage{stfloats}
\usepackage{url}
\usepackage{verbatim}
\usepackage{graphicx}
\usepackage{cite}
\usepackage{graphicx}
\usepackage{multirow}
\usepackage{makecell}
\usepackage{subcaption}
\usepackage{float}
\usepackage{pifont}
\usepackage{circledtext}
\usepackage{enumitem}

\usepackage{amsthm}

\usepackage{booktabs}
\usepackage[switch]{lineno}
\usepackage{tabularx}
\usepackage[table]{xcolor}
\usepackage{nicematrix}
\usepackage{amsthm}
\newtheorem{theorem}{Theorem}

\theoremstyle{definition}

\theoremstyle{remark}

\AtBeginDocument{%
  \fontdimen2\font=0.21em 
  \fontdimen3\font=0.1em  
  \fontdimen4\font=0.06em  
}

\begin{document}

\title{
Adaptive Dual-Weighting Framework for Federated Learning via Out-of-Distribution Detection}

\author{
    Zhiwei~Ling,
    Hailiang~Zhao,~\IEEEmembership{Member,~IEEE},
    Chao~Zhang,
    Xiang~Ao,
    Ziqi~Wang,
    Cheng~Zhang,
    Zhen~Qin,
    Xinkui~Zhao,
    Kingsum~Chow,
    Yuanqing~Wu,~\IEEEmembership{Senior~Member,~IEEE,} and
    MengChu~Zhou,~\IEEEmembership{Fellow,~IEEE}
\thanks{
    Zhiwei Ling, Hailiang Zhao, Chao Zhang, Ziqi Wang, Zhen Qin, Xinkui Zhao, Kingsum Chow, and Hailiang Zhao are with the School of Software Technology, Zhejiang University, and with Zhejiang Key Laboratory of Digital-Intelligence Service Technology (e-mails: \{zwling, hliangzhao, 22351272, wangziqi0312, zhenqin, zhaoxinkui, kingsum.chow\}@zju.edu.cn.).
}
\thanks{
    Cheng Zhang is the School of Information Technology and Artificial Intelligence, Zhejiang University of Finance and Economics, China (e-mail: zhangcheng@zufe.edu.cn).
}
\thanks{
    Xiang Ao is with the College of Computer Science and Technology, Zhejiang University (e-mails: aoxiangyx@zju.edu.cn).
}
\thanks{
    Yuanqing Wu is with the School of Intelligent Systems Engineering, Sun Yat-sen University, Shenzhen 518107, China (e-mail: yqwuzju@163.com).
}
\thanks{
    MengChu Zhou is with the Department of Electrical and Computer Engineering, New Jersey Institute of Technology, Newark, NJ 07102, USA (e-mail: zhou@njit.edu).
}
}


\maketitle

\begin{abstract}
Federated Learning (FL) enables collaborative model training across large-scale distributed service nodes while preserving data privacy, making it a cornerstone of intelligent service systems in edge-cloud environments. 
However, in real-world service-oriented deployments, data generated by heterogeneous users, devices, and application scenarios are inherently non-IID. 
This severe data heterogeneity critically undermines the convergence stability, generalization ability, and ultimately the quality of service delivered by the global model.
To address this challenge, we propose \textsc{FLood}, a novel FL framework inspired by out-of-distribution (OOD) detection. 
\textsc{FLood} dynamically counteracts the adverse effects of heterogeneity through a dual-weighting mechanism that jointly governs local training and global aggregation. 
At the client level, it adaptively reweights the supervised loss by upweighting pseudo-OOD samples, thereby encouraging more robust learning from distributionally misaligned or challenging data. 
At the server level, it refines model aggregation by weighting client contributions according to their OOD confidence scores, prioritizing updates from clients with higher in-distribution consistency and enhancing the global model’s robustness and convergence stability.
Extensive experiments across multiple benchmarks under diverse non-IID settings demonstrate that \textsc{FLood} consistently outperforms state-of-the-art FL methods in both accuracy and generalization. 
Furthermore, \textsc{FLood} functions as an orthogonal plug-in module: it seamlessly integrates with existing FL algorithms to boost their performance under heterogeneity without modifying their core optimization logic.
These properties make \textsc{FLood} a practical and scalable solution for deploying reliable intelligent services in real-world federated environments.
\end{abstract}


\section{Introduction}
\label{sec:introduction}

Federated Learning (FL) has emerged as a transformative paradigm for service-oriented distributed machine learning, enabling collaborative model training across large-scale decentralized clients while preserving data privacy~\cite{fedentropy,feddistil,tsc1}. 
By facilitating continuous optimization of intelligent models over heterogeneous service nodes without centralizing raw data, FL establishes a foundational infrastructure for next-generation data-driven services. 
Its efficacy has been widely validated across diverse service computing domains, including autonomous driving~\cite{driving1,driving2}, intelligent financial systems~\cite{financial1,financial2}, and smart healthcare platforms~\cite{health1,health2}. 

In a typical FL-enabled service workflow, each client receives a global service model from the orchestrating server, performs local training on private service data, and uploads encrypted model updates. 
The server subsequently aggregates these updates to refine the global model for the next communication round. 
This decentralized design ensures that sensitive user data remains on local devices, thereby satisfying stringent privacy requirements and regulatory compliance standards~\cite{tsc2,tsc3,tsc4}. Consequently, FL provides a scalable and privacy-preserving foundation for deploying intelligent services at scale.

\begin{figure}[t]
    \centering 
    \includegraphics[width=0.46\textwidth]{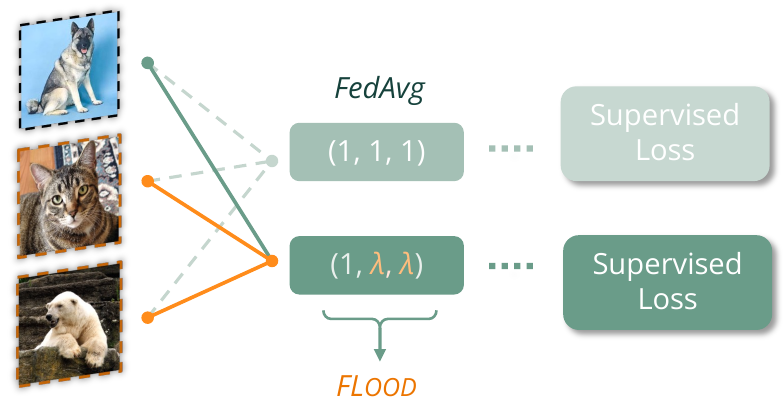}
    \caption{Key difference between \textsc{FLood} and \textsc{FedAvg} on the client side. The orange border highlights pseudo-OOD samples, while $\lambda$ denotes the reallocated weights assigned to these samples in the supervised loss.}
    \label{diff}
\end{figure}

Despite its promise, the practical deployment of FL in real-world service systems is hindered by a fundamental challenge: \textit{data heterogeneity}. 
Service data generated across diverse users, devices, and operational environments inherently exhibits non-independent and identically distributed (non-IID) characteristics~\cite{fedavg, feddecorr}. 
Such distributional discrepancies often induce significant bias in locally trained models. 
When these biased updates are naively aggregated, e.g., via uniform averaging, the resulting global model suffers from unstable convergence, degraded accuracy, and compromised service reliability. 
This issue constitutes a critical bottleneck in realizing scalable, robust, and high-quality federated learning–based intelligent services \cite{asc}.
To mitigate this problem, numerous approaches have been proposed, broadly falling into two categories: server-side model adjustment methods~\cite{FedAvgM,feddisco,fedftg} and client-side training correction techniques~\cite{fedprox,scaffold,moon}. 
Although these methods have demonstrated notable improvements, they primarily operate at the level of model parameters or gradients. 
Consequently, they tend to overlook a more fundamental cause of performance degradation: \textit{the intrinsic mismatch among client data distributions}.

In this work, we introduce a novel perspective on data heterogeneity by reframing it through the lens of out-of-distribution (OOD) detection. 
OOD detection refers to the ability of deep neural networks (DNNs) to identify inputs that deviate significantly from the training distribution~\cite{MSP,ODIN,Energy}. 
While OOD detection has been extensively explored for uncertainty quantification in centralized learning settings, we show that OOD confidence scores also serve as a rich, interpretable, and practically actionable signal for characterizing distributional divergence in federated environments.
Specifically, even though a client’s local dataset may appear internally consistent, certain samples often yield low prediction confidence under the current global model. 
These low-confidence instances can be interpreted as \textit{pseudo-OOD} samples, which quantitatively reflect the degree to which the local data diverges from the global data manifold. 
Importantly, this interpretation transforms data heterogeneity from an unavoidable obstacle into a \textit{measurable and exploitable property} that can guide both local training dynamics and global aggregation policies.

Building upon this insight, we propose \textsc{FLood}, a novel federated learning framework that explicitly and systematically addresses data heterogeneity through a dual-weighting strategy informed by OOD signals. 
As illustrated in Figure~\ref{diff}, in contrast to \textsc{FedAvg}~\cite{fedavg}, \textsc{FLood} enhances local training by assigning higher loss weights to pseudo-OOD samples, thereby encouraging local models to focus more effectively on challenging and distributionally misaligned data. 
On the server side, \textsc{FLood} dynamically modulates client aggregation weights based on their aggregate OOD confidence scores, preferentially incorporating updates from clients whose local data exhibits stronger alignment with the evolving global distribution.
The main contributions of this work are summarized as follows:
\begin{itemize}
    \item We offer a principled reinterpretation of data heterogeneity in FL through the framework of OOD detection, uncovering interpretable signals that facilitate improved cross-client generalization.
    
    \item We design \textsc{FLood}, a federated learning framework that incorporates OOD-aware sample weighting during local training and OOD-guided client reweighting during global aggregation, thereby effectively mitigating the adverse effects of non-IID data.
    
    \item We conduct extensive experiments across multiple benchmark datasets and diverse non-IID configurations, demonstrating that \textsc{FLood} consistently outperforms state-of-the-art FL methods by significant margins.
\end{itemize}

The remainder of this paper is organized as follows. 
Section~\ref{sec:related_work} reviews related work in federated learning and out-of-distribution detection. Section~\ref{sec:preliminaries} introduces the necessary background concepts and problem formulation. 
Section~\ref{sec:approach} details the design of \textsc{FLood}, followed by comprehensive experimental evaluation in Section~\ref{sec:experiment}. 
Finally, Section~\ref{sec:conclusion} summarizes the paper and highlights the main findings.
\section{Related Work}
\label{sec:related_work}

\subsection{Federated Learning}
FL has emerged as a groundbreaking paradigm for distributed machine learning \cite{fedcross,tsc5,tsc6}, yet its effectiveness is often hindered by the challenges posed by non-\textit{i.i.d.} data. 
To address data heterogeneity, existing methods can be broadly categorized into two main directions: \textit{client-side local training correction} and \textit{server-side global model adjustment}.

Compared with the vanilla FL framework (\textsc{FedAvg} \cite{fedavg}), client-side local training correction methods aim to alleviate the inherent biases induced by non-\textit{i.i.d.} data by redefining local optimization objectives. 
For instance, \textsc{FedProx} \cite{fedprox} introduces a proximal regularization term to constrain the divergence between local and global models, thereby facilitating more stable aggregation. 
Similarly, \textsc{MOON} \cite{moon} further improves local training consistency by incorporating a contrastive learning objective to align local and global representations. 
To combat dimensional collapse caused by data heterogeneity, \textsc{FedDecorr} \cite{feddecorr} introduces a decorrelation term to encourage diversity in feature representations. 
\textsc{FedDC} \cite{feddc} explicitly models the discrepancy between local and global models as an additional loss term, thus bridging the parameter gap and improving training stability. 
Moreover, \textsc{FedBR} \cite{fedbr} balances local model outputs to reduce learning bias while preserving the intrinsic characteristics of local data.

Despite these advances, most existing methods generally rely on \textit{auxiliary regularization terms} appended to the standard supervised loss, leaving the loss function itself unchanged.
In contrast, our proposed \textsc{FLood} introduces another perspective by modifying the supervised loss directly. 
By assigning adaptive weights to individual samples based on their OOD confidence scores, \textsc{FLood} prioritizes challenging data points during training. 
This not only addresses the root cause of performance degradation, i.e., data heterogeneity, but also complements existing methods by seamlessly integrating with their regularization-based strategies. 
To the best of our knowledge, \textsc{FLood} is the first framework that incorporates OOD-guided adaptive loss reweighting at the client level.

Global model adjustment methods aim to enhance the global model by introducing techniques such as fine-tuning, knowledge distillation, and dynamic aggregation weight modification, which provide more effective alternatives to traditional direct aggregation strategies. 
For example, \textsc{FedAvgM} \cite{FedAvgM} incorporates server momentum to smooth global updates and reduce fluctuations caused by inconsistent local models. 
Addressing the issue of unequal client data contributions, \textsc{FedNova} \cite{fednova} normalizes aggregation weights to balance the integration of local updates, ensuring fairness and stability in the global process. 
\textsc{FedDisco} \cite{feddisco} further improves aggregation by jointly considering client discrepancy measures and data volumes, achieving tighter error bounds with rigorous theoretical guarantees.
Beyond aggregation strategies, \textsc{FedFTG} \cite{fedftg} employs a data-free knowledge distillation mechanism to fine-tune the global model, enabling knowledge transfer from local models without raw data. 
Additionally, \textsc{FedMR} \cite{fedmr} introduces a layer-wise shuffling strategy for global model parameters during updates, which encourages convergence toward flatter optima and improves robustness.

Although these methods enhance global model aggregation and generalization performance, they predominantly focus on \textit{post-aggregation enhancements}, while overlooking the upstream challenges induced by data heterogeneity and biased local training.
In contrast, \textsc{FLood} explicitly targets these root causes through a dual-weighting strategy that integrates client-side loss reweighting and server-side aggregation adjustment. 
This enables \textsc{FLood} to holistically tackle performance degradation caused by non-\textit{i.i.d.} data and bridge the gap between local training correction and global model adjustment in FL.

\subsection{Out-of-Distribution Detection}
DNN models serve as the core intelligence engines of modern data-driven service systems and typically achieve strong performance on training data as well as in-distribution (ID) test samples \cite{haitao}. 
However, in real-world service-oriented deployments, these models are continuously exposed to a wide spectrum of OOD inputs arising from dynamic user behaviors, diverse service environments, and evolving data sources \cite{cadref}.
Owing to their inherent limitations in recognizing such inputs, misclassifying OOD samples may trigger critical decision errors and incur uncontrollable risks in high-stakes service scenarios.
The primary objective of OOD detection is therefore to endow models with the capability to accurately identify and reject OOD inputs before they adversely affect service accuracy, system reliability, and overall quality of service.

Among existing approaches, post-hoc scoring methods are particularly appealing due to their simplicity, computational efficiency, and strong compatibility with pre-trained models. 
These methods avoid modifying the model architecture and instead rely on model outputs, such as logits or intermediate features, to perform OOD classification.
As a result, they can be seamlessly integrated into existing pipelines with minimal additional computational overhead. 
Along this line, a variety of post-hoc OOD scoring methods have been proposed to quantify predictive confidence, which can be broadly categorized into \textit{logit-based} and \textit{feature-based} approaches.
As one of the earliest and most representative logit-based methods, Hendrycks \textit{et al.} \cite{MSP} adopt the maximum softmax probability of the logits as the OOD scoring function. 
Following this line, Energy \cite{Energy} further constructs OOD scores based on the energy function derived from logits, providing a more expressive confidence measure.
For feature-based methods, ReAct \cite{ReAct} mitigates the influence of outlier activations by truncating feature values that exceed a predefined threshold, thereby suppressing abnormal responses in hidden representations. 
ASH \cite{ASH} further proposes a simple feature suppression strategy that removes a large portion of sample features while selectively adjusting the remaining ones.
\section{Preliminaries}
\label{sec:preliminaries}

\subsection{FL Problem Setup}
FL involves a set of $N$ clients, denoted as ${\mathcal{C}_1, \mathcal{C}_2, \ldots, \mathcal{C}_N}$, where each client $\mathcal{C}_k$ holds a local dataset $\mathcal{D}_k$. 
The objective of FL is to collaboratively optimize a global model $\theta$ while preserving the privacy of client data. 
This objective is achieved by minimizing the overall loss across all clients, which is formulated as
\begin{equation}
\arg \min_\theta \left\{\mathcal{F}(\theta) =\sum_{k=1}^{N} p_k \mathcal{L}_k(\theta)\right\},  
\end{equation}

where $\mathcal{L}_k(\theta)$ represents the local empirical loss of client $\mathcal{C}_k$, defined as
\begin{equation}
\mathcal{L}_k(\theta) := \frac{1}{n_k} \sum_{i=1}^{n_k} \Phi(\theta; x_i, y_i).
\end{equation}

Here, $(x_i, y_i)$ represents the $i$-th training sample on client $\mathcal{C}_k$, $n_k$ is the number of samples on the client, and $\Phi(\theta; x, y)$ denotes the per-sample loss function. 
The aggregation weight $p_k$ is proportional to the size of the client’s dataset, given by
\begin{equation}
    p_k = \frac{n_k}{\sum_{i=1}^{N} n_i}.
\end{equation}
 
During the $t$-th communication round, a subset of clients $\mathcal{C}^t \subseteq { \mathcal{C}_1, \mathcal{C}_2, \ldots, \mathcal{C}_N }$ is selected to participate in the training process.
Each selected client $\mathcal{C}_k \in \mathcal{C}^t$ receives the current global model $\theta^t$ from the server, performs several steps of local optimization on its private dataset $\mathcal{D}_k$, and uploads the updated local model $\theta_k^t$ to the server.
The server then aggregates the received local models to update the global model as

\begin{equation}
    \theta^{t+1} = \sum_{k: \mathcal{C}_k \in \mathcal{C}^t} p_k \theta_k^t.
\end{equation}
This process is repeated over multiple communication rounds until the global model $\theta$ converges.

\subsection{Post-Hoc OOD Score}

Let $\mathcal{D}_{\text{id}}$ and $\mathcal{D}_{\text{ood}}$ denote the ID dataset and the OOD dataset, respectively.
The goal of OOD detection is to determine whether a given sample $(x, y)$ belongs to $\mathcal{D}_{\text{id}}$ or $\mathcal{D}_{\text{ood}}$.
For a given DNN model $\theta$, post-hoc OOD detection techniques employ a heuristic scoring function $\textsc{S}(\cdot; \cdot)$ to assign an OOD confidence score to each input sample $(x, y)$.
Representative methods such as MSP \cite{MSP} and Energy \cite{Energy} adopt different scoring strategies.
Specifically, MSP computes the maximum softmax probability of the model output, whereas Energy evaluates the ``energy'' value of the predicted probability distribution.
These scores serve as effective indicators of whether a test sample is consistent with the training distribution. 
In practical applications, samples are classified as either ID or OOD according to the following decision criterion:

\begin{equation}
(x, y) \sim
    \begin{cases}
    \mathcal{D}_{\text{ood}}, & \text{if } \textsc{S}(\theta; x) \leq \tau, \\
    \mathcal{D}_{\text{id}}, & \text{if } \textsc{S}(\theta; x) > \tau,
    \end{cases}
    \label{ood_objective}
\end{equation}
where $\tau$ is a predefined threshold that separates $\mathcal{D}{\text{id}}$ from $\mathcal{D}{\text{ood}}$. 
As described in Eq. \eqref{ood_objective}, samples with confidence scores above $\tau$ are classified as ID, whereas those with scores below $\tau$ are identified as OOD. This threshold-based decision rule is computationally efficient and introduces negligible deployment overhead, making post-hoc OOD detection methods particularly suitable for resource-constrained scenarios or as lightweight plug-in modules for pre-trained models in real-world systems.

\section{The FLood Framework}
\label{sec:approach}

In this section, we present \textsc{FLood} (\underline{F}ederated \underline{L}earning via \underline{OOD} Detection), a novel framework that enhances FL under non-IID data settings by integrating OOD detection into both client-side training and server-side aggregation. 
The core idea of \textsc{FLood} is to dynamically adjust the importance of individual samples during local optimization and the reliability of different clients during global aggregation, based on their estimated distributional alignment with the evolving global model. 
By leveraging OOD awareness as an adaptive control signal, \textsc{FLood} establishes a principled mechanism to mitigate the adverse effects of data heterogeneity.

\begin{figure*}[htbp]
    \centering
    \includegraphics[width=7.1in]{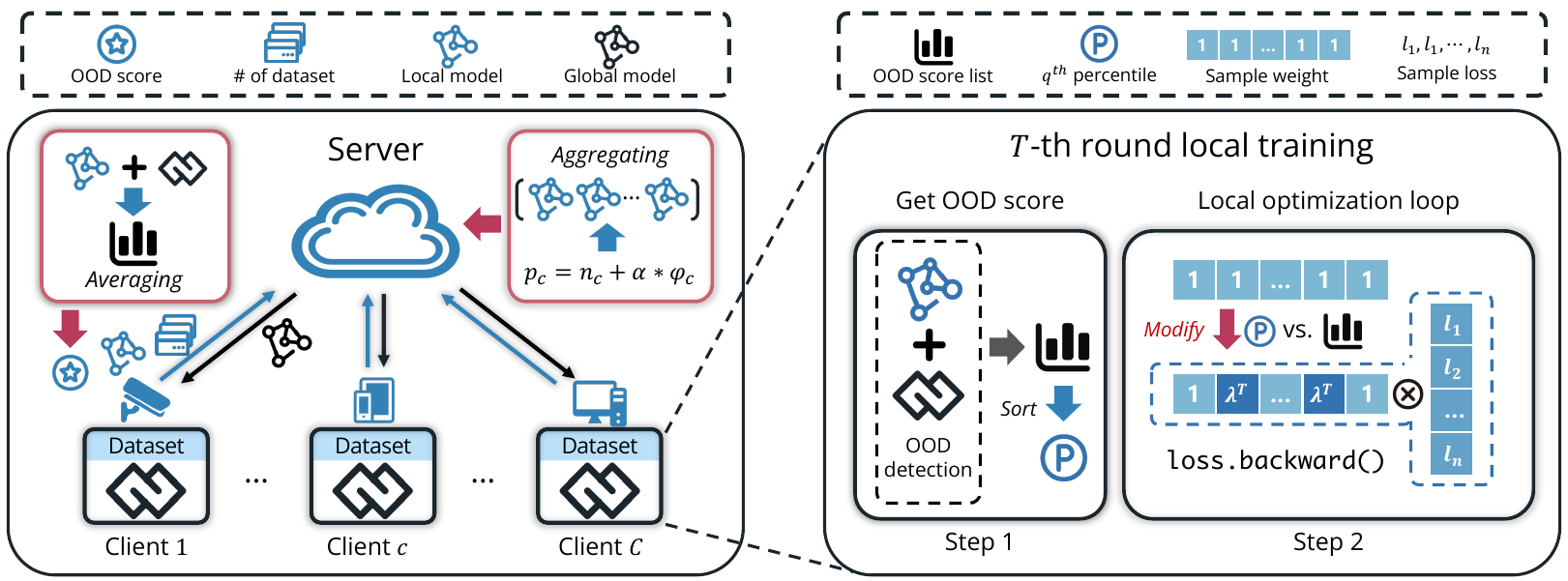}
    \caption{Framework and workflow of \textsc{FLood}. The right illustrates the client-side training process, where OOD detection is used to adjust sample weights in the supervised loss. The left demonstrates the server-side process, where the server dynamically reallocates aggregation weights based on client-provided information.}
    \label{Overview}
\end{figure*}

The architecture and workflow of \textsc{FLood} are illustrated in Figure \ref{Overview}. 
It introduces two tightly coupled components, namely \textit{Adaptive Sample Weighting} (Section \ref{c1}) and \textit{Dynamic Aggregation Correction} (Section \ref{c2}), which together form a dual-weighting strategy.
Specifically, the former emphasizes locally low-confidence samples during client-side optimization, while the latter adjusts aggregation weights at the server according to the model confidence on the local dataset.

\subsection{Overview}
Figure \ref{Overview} illustrates the framework and workflow of \textsc{FLood}. 
Unlike conventional FL methods, \textsc{FLood} adaptively modulates the training process by assigning dynamic weights to individual samples based on their OOD confidence scores during client-side optimization.
Concurrently, it refines the server-side aggregation by adjusting the aggregation weights based on the OOD information reported by each client. 
This dual-weighting strategy jointly enhances the robustness and generalization capability of the global model.
Each training round $t$ of \textsc{FLood} consists of the following steps:
\begin{enumerate}
    \item 
    \textit{Client Selection and Model Broadcasting:}
    At the beginning of the $t$-th training round, the server randomly selects a subset of clients $\mathcal{C}^t$ and dispatches the current global model $\theta^t$ to each selected client. 
    This step follows the standard protocol of conventional FL frameworks.
    
    \item 
    \textit{OOD-Aware Local Training:}
    Upon receiving the global model $\theta^t$, each client performs OOD detection on its local dataset to assess sample-level confidence scores, thereby partitioning its data into {genuine-ID} (inlier) and {pseudo-OOD} (outlier) subsets based on a predefined threshold. 
    During the local training phase, \textsc{FLood} assigns higher loss weights to pseudo-OOD samples, thereby prioritizing learning from these challenging data points.
    Note that as the model evolves, the classification of samples may dynamically change over training iterations.
    
    \item 
    \textit{OOD Confidence Aggregation and Upload:}
    After completing local training, each client aggregates the OOD confidence scores of its local dataset into a single scalar $\phi_c$. 
    Together with the updated local model $\theta_c^t$ and the dataset size $n_c$, the aggregated confidence score $\phi_c$ is uploaded to the server. 
    The inclusion of $\phi_c$ allows the server to make informed adjustments to the global aggregation weights.
    
    \item 
    \textit{Confidence-Guided Global Aggregation:}
    Based on the received local models, dataset sizes, and aggregated OOD confidence scores, the server computes new aggregation weights for updating the global model. 
    Clients with higher confidence scores $\phi_c$ are assigned larger aggregation weights and thus contribute more significantly to the global update $\theta^{t+1}$, enhancing robustness and generalization under heterogeneous data distributions.
    
\end{enumerate}

\begin{algorithm}[tb]
    \caption{\textsc{FLood} (the server side)}
    \label{a1}
    \textbf{Input}: $R$: number of communication rounds; $N$: total number of clients; $M$: number of clients selected per round. \\
    \textbf{Output}: $\theta^T$: final global model after $T$ rounds.

    \begin{algorithmic}[1]
        \STATE Initialize global model: $\theta^0$ \label{line:init}
        \FOR{$t = 0$ to $R-1$}
            \STATE Randomly select $M$ clients: $\mathcal{C}^t \subseteq \mathcal{C}$ \label{line:select}
            \STATE Broadcast $\theta^t$ to all clients in $\mathcal{C}^t$ \label{line:broadcast}
            \FOR{each client $c \in \mathcal{C}^t$ \textbf{in parallel}}
                \STATE \textcolor{blue}{$\theta_c^t, n_c, \phi_c \leftarrow$ \textsc{LocalUpdate}($\theta^t, t$)} \label{line:local_update}
            \ENDFOR
            \STATE \textcolor{blue}{Compute client weights: $p_c = f(\phi_c)$ via ~\eqref{server_weight}} \label{line:weight}
            \STATE Aggregate: $\theta^{t+1} \leftarrow \sum_{c \in \mathcal{C}^t} p_c \theta_c^t$ \label{line:aggregate}
        \ENDFOR
        \RETURN $\theta^T$ \label{line:return}
    \end{algorithmic}
\end{algorithm}

Algorithm~\ref{a1} provides a high-level summary of the server-side logic, where blue-colored blocks indicate the unique aspects introduced by \textsc{FLood} compared to standard FL frameworks like \textsc{FedAvg}~\cite{fedavg}.
Line~\ref{line:init} initializes the global model $\theta^0$, and Lines~\ref{line:select}-\ref{line:aggregate} describe the core steps within each communication round.
In each round, the server randomly selects $M$ clients to form the active set $\mathcal{C}^t$ as described in Line~\ref{line:select}, and then broadcasts the current global model $\theta^t$ to these clients as in Line~\ref{line:broadcast}.
Lines~5-7 (corresponding to the parallel execution block) depict the client-side updates, where each active client obtains its local model $\theta_c^t$, dataset size $n_c$, and aggregated OOD confidence score $\phi_c$ (Line~\ref{line:local_update}).
In Line~\ref{line:weight}, the server computes the aggregation weight $p_c$ for each client based on $\phi_c$ using ~\eqref{server_weight}.
Line~\ref{line:aggregate} aggregates all local models to produce the updated global model $\theta^{t+1}$.
After $R$ rounds, Line~\ref{line:return} outputs the final global model $\theta^T$.

\subsection{Adaptive Sample Weighting}
\label{c1}

To address intra-client data heterogeneity, \textsc{FLood} leverages post-hoc OOD detection techniques to uncover samples that deviate from the expected data distribution. 
Specifically, \textsc{FLood} computes an OOD confidence score for each sample, which reflects the model’s confidence in its prediction for the primary task. 
Higher scores indicate greater predictive certainty, whereas lower scores imply higher uncertainty. 
By applying a predefined threshold, the local dataset is partitioned into two categories: genuine in-distribution (\textit{genuine-ID}) samples and pseudo out-of-distribution (\textit{pseudo-OOD}) samples. 
Intuitively, pseudo-OOD samples, which exhibit higher uncertainty, require targeted learning emphasis, while genuine-ID samples with reliable predictions require less intervention.

\begin{algorithm}[h]
    \caption{\textsc{LocalUpdate} of \textsc{FLood} (on client $c$)}
    \label{a2}
    \textbf{Input}: $\theta^t$: global model at round $t$; $E$: number of local epochs; $\mathcal{D}_c$: client's local dataset; $\eta$: learning rate; $T$: stopping round for weight growth. \\
    \textbf{Output}: $\theta_c^t$: updated local model; $n_c$: number of samples; $\phi_c$: aggregated OOD score.

    \begin{algorithmic}[1]
        \STATE Initialize local model: $\theta_c^t \leftarrow \theta^t$ \label{line:local_init}
        \FOR{$e = 0$ to $E-1$} \label{line:epoch_start}
            \FOR{mini-batch $(X_b, Y_b) \sim \mathcal{D}_c$} \label{line:batch_start}
                \STATE Compute base losses: $\texttt{Loss} \leftarrow \Phi(\theta_c^t; X_b, Y_b)$ \label{line:base_loss}
                \STATE \textcolor{blue}{Compute \texttt{Score}, $\tau$ via Eqs.~\eqref{score} and~\eqref{score&thre}} \label{line:compute_score}
                \STATE \textcolor{blue}{Compute \texttt{Mask} via ~\eqref{mask}} \label{line:compute_mask}
                \STATE \textcolor{blue}{Weighted loss: $\texttt{Loss} \leftarrow \textsc{Mean}(\texttt{Loss} \circ \texttt{Mask})$} \label{line:weighted_loss}
                \STATE Update model: $\theta_c^t \leftarrow \theta_c^t - \eta \nabla_{\theta_c^t} \texttt{Loss}$ \label{line:update}
            \ENDFOR \label{line:batch_end}
        \ENDFOR \label{line:epoch_end}
        \STATE \textcolor{blue}{Aggregate OOD score: $\phi_c \leftarrow \frac{1}{|\mathcal{D}_c|} \sum_{x \in \mathcal{D}_c} \textsc{Score}_x$} \label{line:agg_score}
        \RETURN $\theta_c^t, |\mathcal{D}_c|, \phi_c$ \label{line:local_return}
    \end{algorithmic}
\end{algorithm}

Algorithm~\ref{a2} outlines the \textsc{LocalUpdate} procedure executed on each client, with blue-colored lines indicating the integration of our adaptive weighting mechanism.
Line~\ref{line:local_init} initializes the client model $\theta_c^t$ using the global model received from the server.
Lines~\ref{line:epoch_start}--\ref{line:epoch_end} describe the local optimization process over $E$ epochs.
During each epoch, the client iterates over mini-batches drawn from the local dataset $\mathcal{D}_c$, as specified in Line~\ref{line:batch_start}.
Line~\ref{line:base_loss} computes the standard supervised loss for each sample in the current mini-batch, denoted as $\texttt{Loss} = [l_1, l_2, \ldots]$.
Before the model update, \textsc{FLood} computes a confidence score for each input sample $x$ in the mini-batch $(X_b, Y_b)$:
\begin{align}
    \textsc{Score}_x := \textsc{S}(\theta_c^t, x),
    \label{score}
\end{align}
where $\textsc{S}(\cdot; \cdot)$ denotes a chosen out-of-distribution (OOD) scoring function, such as energy-based scoring~\cite{Energy} or maximum softmax probability (MSP)~\cite{MSP}.
The vector of scores across the batch is denoted as $\texttt{Score} = [\textsc{Score}_1, \textsc{Score}_2, \ldots]$.
Subsequently, a dynamic threshold $\tau$ is determined as the $q$-th percentile of the $\texttt{Score}$ vector:
\begin{equation}
    \tau = [\texttt{Score}]_{q\%}.
    \label{score&thre}
\end{equation}
This threshold enables the identification of pseudo-OOD samples, which are then assigned higher loss weights through the masking operation in Line~\ref{line:compute_mask}.
The reweighted loss is computed in Line~\ref{line:weighted_loss}, and the local model is updated accordingly in Line~\ref{line:update}.
Finally, after completing all local epochs, the client aggregates the OOD scores over its entire dataset to obtain $\phi_c$, as shown in Line~\ref{line:agg_score}, and returns the updated model, dataset size, and OOD confidence score in Line~\ref{line:local_return}.

Using this threshold, we classify each sample as either genuine-ID ($\textsc{Score}_x \geq \tau$) or pseudo-OOD ($\textsc{Score}_x < \tau$).
To prioritize learning from pseudo-OOD samples, \textsc{FLood} employs an adaptive weighting mechanism. 
Each sample $(x, y)$ is assigned a weight \texttt{Mask}, defined as:
\begin{align}
    \texttt{Mask}_x^t = \lambda^t \cdot \mathbb{I}(\textsc{Score}_x < \tau) + \mathbb{I}(\textsc{Score}_x \geq \tau),
    \label{mask} 
\end{align}
where $\mathbb{I}(\cdot)$ is the indicator function.
For samples with confidence scores greater than the threshold $\tau$, their weights in 
\texttt{Mask} are set to 1; otherwise, their weights are set to $\lambda^t$.

\begin{figure}[htbp]
    \centering    
    \begin{subfigure}{0.36\textwidth}
        \centering
        \includegraphics[width=\linewidth]
        {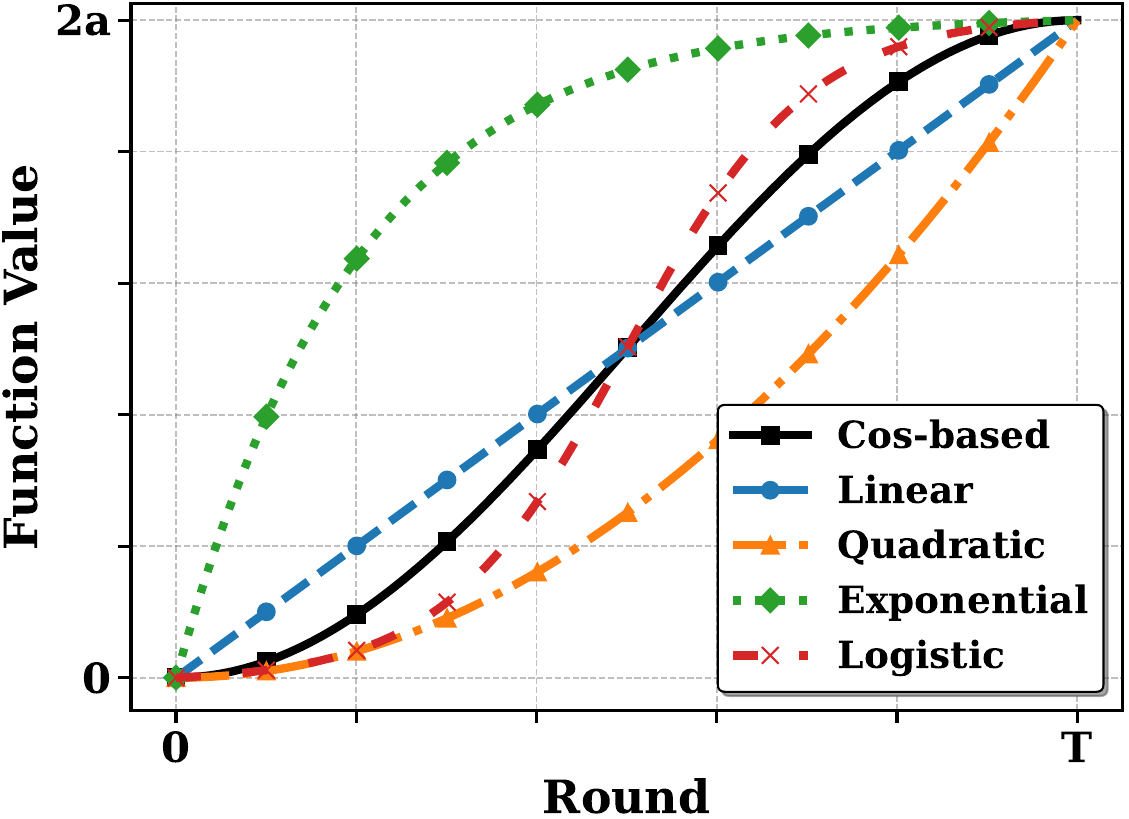}
        \caption{Scheduling functions.}
        \label{function}
    \end{subfigure}
    \hfill \\
    \vspace{0.05in}
    \begin{subfigure}{0.36\textwidth}
        \centering
        \includegraphics[width=\linewidth]{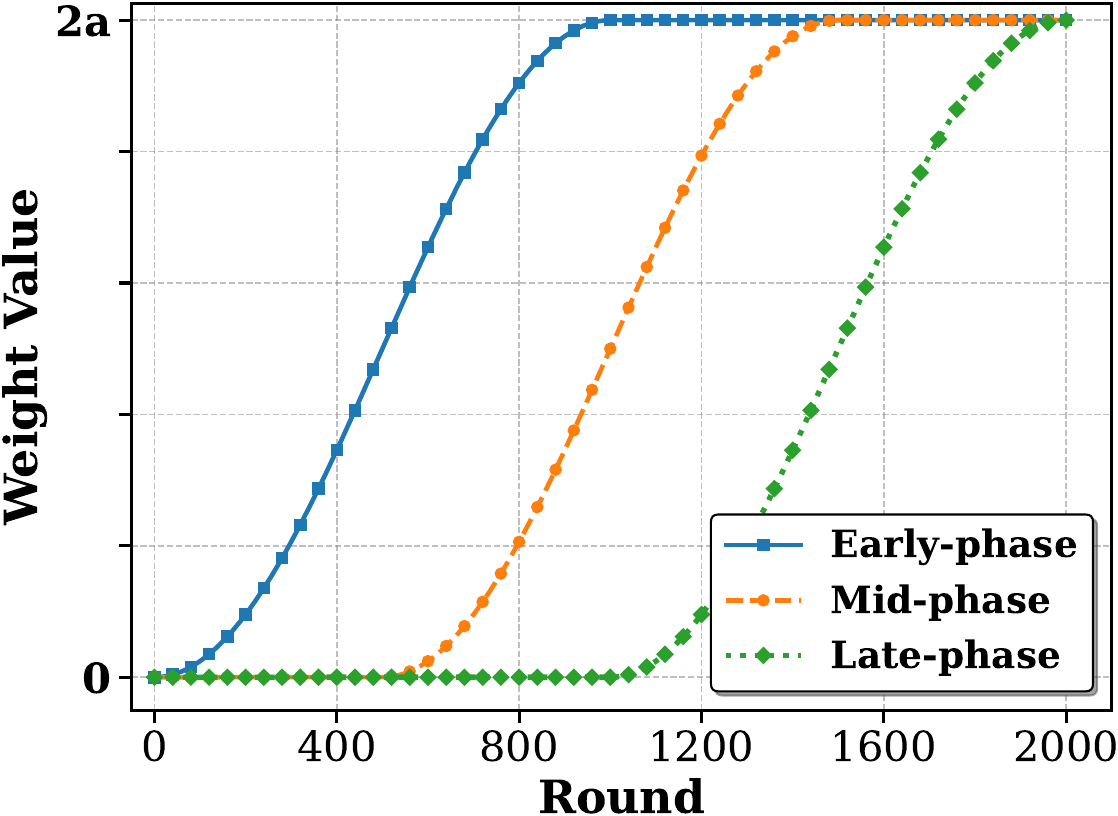}
        \caption{Increasing phases.}
        \label{phase}
    \end{subfigure}
    \caption{Illustration of the weight adjustment curve in \textsc{FLood}. 
    (a) Comparison of different weight scheduling functions, including cosine-based, linear, quadratic, exponential, and logistic schedules. 
    (b) Weight growth under different starting phases.
    }
\end{figure}

To better facilitate the learning of these pseudo-OOD samples, we adopt a progressively increasing weight schedule during training.
As the model becomes more capable, the growth of these weights is halted at round $T$ to prevent adverse effects on the learning of genuine-ID samples.
Moreover, since the model’s OOD detection capability is unreliable in the early stages, excessively rapid weight increases may cause training instability. 
As the weight growth approaches its stopping point, the rate of increase should also diminish.
To achieve this behavior, \textsc{FLood} employs a cosine function to regulate the growth of the weights, as defined by the following formula:
\begin{equation}
   \lambda^t = a \cdot \left(1 - \cos \frac{\pi t}{T} \right),
\end{equation}
where $a$ controls the maximum amplification factor and $T$ denotes the round at which $\lambda^t$ reaches its plateau.
This scheduling strategy mitigates early-stage instability arising from immature OOD estimates, while progressively strengthening the influence of pseudo-OOD samples as the model becomes increasingly reliable over time.

Beyond the cosine schedule, we further investigated several alternative weighting functions and growth periods to examine how different dynamics influence the effectiveness. 
The candidate functions are defined as follows:
\begin{itemize}
    \item \textbf{Linear:} $\lambda^t = 2a \cdot \frac{t}{T}$
    \item \textbf{Quadratic:} $\lambda^t = 2a \cdot (\frac{t}{T})^2$
    \item \textbf{Exponential:} $\lambda^t = 2a \cdot \frac{1-\exp(-kt)}{1-\exp(-kT)}$
    \item \textbf{Logistic:} $\lambda^t = 2a \cdot \frac{\sigma(\alpha(t-T/2))-\sigma(-\alpha T/2)}{\sigma(\alpha T / 2)-\sigma(-\alpha T/2)}$, where $\sigma(\cdot)$ denotes the logistic sigmoid function
\end{itemize}
Figure \ref{function} visualizes the growth patterns of these scheduling functions, illustrating their differences in initial growth speed and convergence behavior.
To further analyze the temporal characteristics, we also vary the initiation point of weight growth and consider three configurations: early start, middle start, and late start, as shown in Figure \ref{phase}.

\subsection{Dynamic Aggregation Correction}
\label{c2}

Building on the insights of \textsc{FedDisco} \cite{feddisco} and \textsc{FedNova} \cite{fednova}, \textsc{FLood} introduces a dynamic aggregation mechanism that adjusts client weights according to their OOD confidence scores. 
This design strengthens the robustness of the global model by adaptively reflecting the quality of local updates. 
In contrast, \textsc{FedDisco} relies on static data discrepancy measures, and \textsc{FedNova} adjusts aggregation weights based on effective iteration counts rather than model reliability.

Upon completing local training, each client $c$ performs OOD detection on its local dataset $\mathcal{D}_c$. 
The OOD score computation is lightweight, as it does not involve model updates, thus minimizing computational overhead. 
For each sample $(x, y)$ in the client’s dataset, an OOD confidence score is obtained using the function $\textsc{S}(\theta^t_c; x)$, where $\theta^t_c$ represents the local model at round $t$. 
The aggregated OOD score $\phi_c$ for client $c$ is then calculated as:
\begin{equation}
    \phi_c = \frac{1}{n_c} \sum_{x \in \mathcal{D}_c} \textsc{S}(\theta^t_c; x),
    \label{phi}
\end{equation}
where $n_c$ denotes the size of the local dataset $\mathcal{D}_c$. 
The scalar value $\phi_c$ is transmitted to the server together with the updated local model $\theta_c^t$ and dataset size $n_c$, introducing negligible communication overhead. 
Moreover, this design preserves data privacy, as $\phi_c$ does not reveal specific information about the class distribution or sensitive characteristics of the dataset.

To incorporate the OOD score $\phi_c$ into the aggregation process, \textsc{FLood} extends the traditional data volume-based weighting scheme. 
Specifically, the adjusted aggregation weight for client $k$ is defined as:
\begin{equation}
    p_c = \textsc{Norm}\Big(\textsc{Norm}(n_c) + \alpha \cdot \textsc{Norm}(\phi_c)\Big),
    \label{server_weight}
\end{equation}
where $\textsc{Norm}(\cdot)$ denotes a normalization operation that scales each component to a comparable range, and $\alpha$ is a tunable parameter that controls the relative influence of the OOD score on the final aggregation weight. 
This formulation ensures that clients with higher OOD scores, indicating greater confidence in their understanding of local data distributions, contribute more significantly to the global model update.

By integrating both data volume and OOD confidence into the aggregation weights, \textsc{FLood} achieves a balance between exploiting large datasets and prioritizing high-quality updates from confident local models. 
Consequently, the global model becomes more robust against data heterogeneity, effectively capturing diverse patterns across different clients while mitigating the risk of overfitting to any particular subset of the data. 
Furthermore, this dynamic weighting scheme allows \textsc{FLood} to adaptively adjust the contribution of each client in response to evolving data distributions and model capabilities. 
As the global model improves, the relative importance of individual client updates naturally shifts, leading to a more resilient and generalizable federated model.

\subsection{Convergence Analysis}
\label{sec:convergence}

We now provide a convergence analysis for \textsc{FLood} under standard assumptions in federated optimization. 
Let $F(\theta) = \sum_{c=1}^N \frac{n_c}{n} \mathcal{L}_c(\theta)$ denote the global objective, where $\mathcal{L}_c(\theta) = \mathbb{E}_{(x,y)\sim\mathcal{D}_c}[\Phi(\theta; x, y)]$ is the true local risk on client $c$, and $n = \sum_{c=1}^N n_c$.

In \textsc{FLood}, each client $c$ optimizes a reweighted surrogate loss at round $t$:
\begin{equation}
    \widetilde{\mathcal{L}}_c^t(\theta) := \frac{1}{n_c} \sum_{i=1}^{n_c} w_{c,i}^t \, \Phi(\theta; x_i, y_i),
\end{equation}
where the sample weights $w_{c,i}^t = \texttt{Mask}_{x_i}^t > 0$ are determined by the OOD confidence scores of the current global model $\theta^t$ (see ~\eqref{mask}). 
Note that $\widetilde{\mathcal{L}}_c^t$ is time-varying due to its dependence on $\theta^t$.
After local training, the server aggregates updates using dynamic weights $p_c^t$ computed from the reported OOD confidence scores $\phi_c^t$ (see~\eqref{server_weight}). 
The global update is:
\begin{equation}
    \theta^{t+1} = \sum_{c \in \mathcal{C}^t} p_c^t \, \theta_c^t,
\end{equation}
where $\theta_c^t$ is the local model returned by client $c$.

To analyze convergence, we make the following assumptions:

\begin{enumerate}[label=(A\arabic*), ref=A\arabic*]
    \item Each $\mathcal{L}_c$ is $L$-smooth, i.e., 
    $\|\nabla \mathcal{L}_c(\theta) - \nabla \mathcal{L}_c(\theta')\| \leq L \|\theta - \theta'\|$ for all $\theta, \theta'$.
    \label{ass:smooth}

    \item The stochastic gradient satisfies 
    $\mathbb{E}_{(x,y)}[\|\nabla \Phi(\theta; x, y) - \nabla \mathcal{L}_c(\theta)\|^2] \leq \sigma^2$ for all $c$ and $\theta$.
    \label{ass:variance}

    \item There exist constants $W_{\min}, W_{\max} > 0$ such that 
    $W_{\min} \leq w_{c,i}^t \leq W_{\max}$ for all $c, i, t$. This holds by design since $\lambda^t \leq 2a$ (Sec.~\ref{c1}).
    \label{ass:weights}

    \item Define the local gradient bias as 
    $b_c^t := \nabla \widetilde{\mathcal{L}}_c^t(\theta^t) - \nabla \mathcal{L}_c(\theta^t)$. 
    We assume $\|b_c^t\| \leq \delta$ for some $\delta \geq 0$ and all $c, t$.
    \label{ass:bias}

    \item Let $q_c = n_c / n$ be the ideal data-proportional weight. 
    The actual aggregation weights $p_c^t$ satisfy
    \begin{equation}
        \left\| \sum_{c \in \mathcal{C}^t} p_c^t \nabla \mathcal{L}_c(\theta^t) - \nabla F(\theta^t) \right\| \leq \epsilon,
    \end{equation}
    for some $\epsilon \geq 0$.
    \label{ass:agg_bias}
\end{enumerate}

Assumptions \ref{ass:bias} and \ref{ass:agg_bias} capture the deviation introduced by \textsc{FLood}'s OOD-aware mechanisms. In practice, well-calibrated OOD scoring and moderate weighting schedules (e.g., cosine annealing) ensure that $\delta$ and $\epsilon$ remain small.
Under these assumptions, we obtain the following convergence guarantee.

\begin{theorem}[Convergence of \textsc{FLood}]
Suppose the learning rate is set to $\eta = \frac{1}{\sqrt{T}}$. Then the iterates $\{\theta^t\}_{t=0}^{T-1}$ generated by \textsc{FLood} satisfy
\begin{equation}
    \frac{1}{T} \sum_{t=0}^{T-1} \mathbb{E} \left[ \left\| \nabla F(\theta^t) \right\|^2 \right] 
    \leq \mathcal{O} \left(\frac{1}{\sqrt{T}}\right) + \mathcal{O} \left((\epsilon + \delta)^2\right),
\end{equation}
where $\epsilon$ and $\delta$ are defined in Assumptions \ref{ass:bias} and \ref{ass:agg_bias}.
\label{thm:main}
\end{theorem}
The proof is provided in the supplementary material.


\section{Experiments}
\label{sec:experiment}
In this section, we implement \textsc{FLood} using PyTorch \cite{Pytorch} to assess the performance.\footnote{Our code is available at \url{https://github.com/LingAndZero/FLood}}
All experiments are conducted on an Ubuntu workstation equipped with 512 GB of RAM, an AMD EPYC 7Y43 CPU, and two NVIDIA GeForce RTX 4090 GPUs.
To systematically evaluate the effectiveness and generality of the proposed framework, we organize our experimental study around four core Research Questions (RQs):

\textit{\textbf{RQ1 (Superiority):}} Does \textsc{FLood} outperform state-of-the-art FL methods across different levels of data heterogeneity?

\textit{\textbf{RQ2 (Scalability):}}
How well does \textsc{FLood} scale under different FL settings in practice?

\textit{\textbf{RQ3 (Compatibility):}}
Can \textsc{FLood} be effectively integrated with existing client-side training correction methods?

\textit{\textbf{RQ4 (Ablation):}}
Do all key components of \textsc{FLood} contribute to its overall performance improvement, and how sensitive is the framework to its hyperparameters?

\renewcommand{\arraystretch}{1.8}
\begin{table}[h!]
\caption{Hyperparameter settings for different FL baselines.}
\label{hyperparameters}
\begin{center}
\begin{tabular}{llc}
    \toprule
    \textbf{Baselines} & \textbf{Hyperparameters} & \textbf{Value} \\
    \midrule
    \textsc{FedAvg}      & None  & -- \\ \hline
    \textsc{FedProx}     & $\mu$:  regularization coefficient   & $\mu = 0.1$ \\
    \textsc{SCAFFOLD}    & None  & -- \\
    \textsc{Moon}        & 
    $ \begin{cases} \mu: \text{regularization coefficient} \\
    \tau: \text{temperature}  \end{cases} $    & $ \begin{cases} \mu = 1.0 \\ \tau = 0.5 \end{cases}$ \\
    \textsc{FedDecorr}   & $\beta$: regularization coefficient & $\beta = 0.1$ \\
    \textsc{FedNP}       & $\lambda$: regularization coefficient & $\lambda = 0.01$ \\
    \textsc{FedAvgM}     & $\rho$: server momentum & $\rho = 0.1$ \\
    \textsc{FedNova}     & None & -- \\
    {\multirow{2}{*}{\textsc{FedDisco}}}    & $ \begin{cases} a: \text{balance coefficient} \\ b: \text{adjustment coefficient} \end{cases}$ & $\begin{cases} a = 0.5 \\ b = 1 \end{cases} $ \\
    \textsc{FedMR}       & None & -- \\
    \midrule
    \textsc{FLood} & $\begin{cases} q: \text{OOD detection threshold}\\ 
    a: \text{amplification coefficient} \\ T: \text{halt round} \\ \alpha: \text{aggregation ratio} \end{cases} $  
    & $ \begin{cases} q =0.7 \\ a= 200 \\ T = 1000 \\ \alpha = 0.5 \end{cases} $ \\
   \bottomrule
\end{tabular}
\end{center}
\end{table}
\subsection{Setup}

\paragraph{Baselines}
We compare \textsc{FLood} against ten state-of-the-art federated learning methods, including the classic baseline \textsc{FedAvg} \cite{fedavg}, five client-side local training correction approaches (\textsc{FedProx} \cite{fedprox}, \textsc{SCAFFOLD} \cite{scaffold}, \textsc{Moon} \cite{moon}, \textsc{FedDecorr} \cite{feddecorr} and \textsc{FedNP} \cite{fednp}, as well as four server-side global model adjustment methods (\textsc{FedAvgM} \cite{FedAvgM}, \textsc{FedNova} \cite{fednova}, \textsc{FedDisco} \cite{feddisco}, and \textsc{FedMR} \cite{fedmr}).
For \textsc{FLood}, we adopt the \textsc{Energy} score \cite{Energy} as the default OOD detection method. 
On the client side, the OOD threshold $q$ is set to $70\%$, the amplification factor $a$ to $200$, and the weight stabilization round $T$ to $1000$. 
On the server side, the aggregation factor $\alpha$ is fixed to $0.5$. 
Detailed hyperparameter configurations for all baseline methods are summarized in Table
\ref{hyperparameters}.

\begin{figure}[htbp]
    \centering
    \begin{subfigure}{0.158\textwidth}
        \centering
        \includegraphics[width=\textwidth]{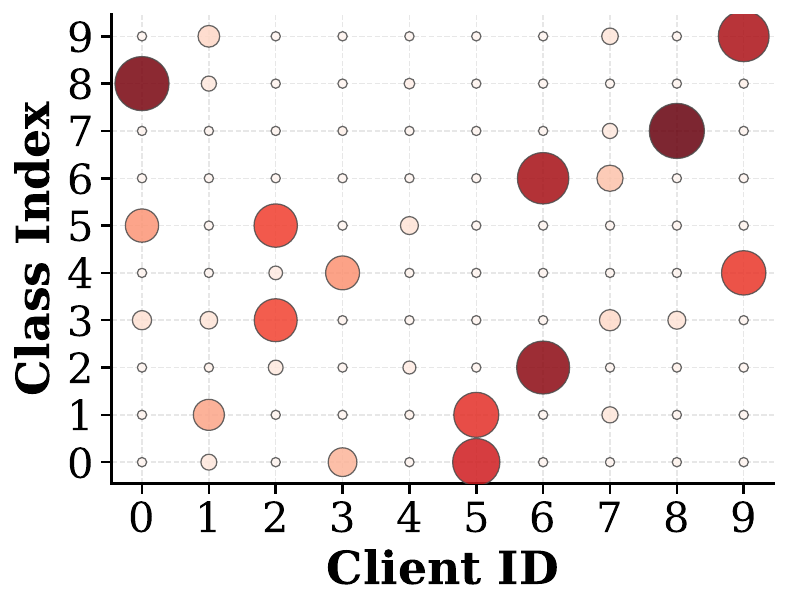}
        \caption{$\text{Dir}(0.1)$}
    \end{subfigure}
    \hfill
    \begin{subfigure}{0.158\textwidth}
        \centering
        \includegraphics[width=\textwidth]{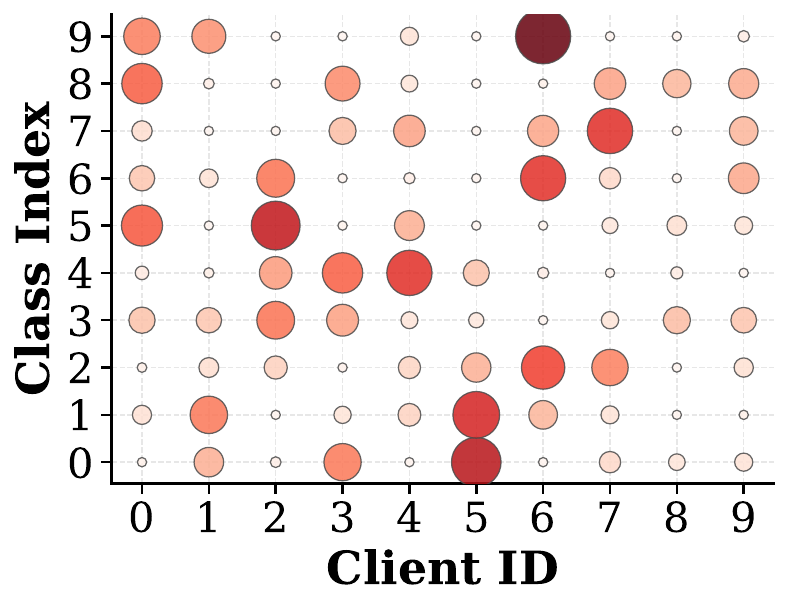}
        \caption{$\text{Dir}(0.5)$}
    \end{subfigure}
    \vspace{0.1in}
    \hfill
    \begin{subfigure}{0.158\textwidth}
        \centering
        \includegraphics[width=\textwidth]{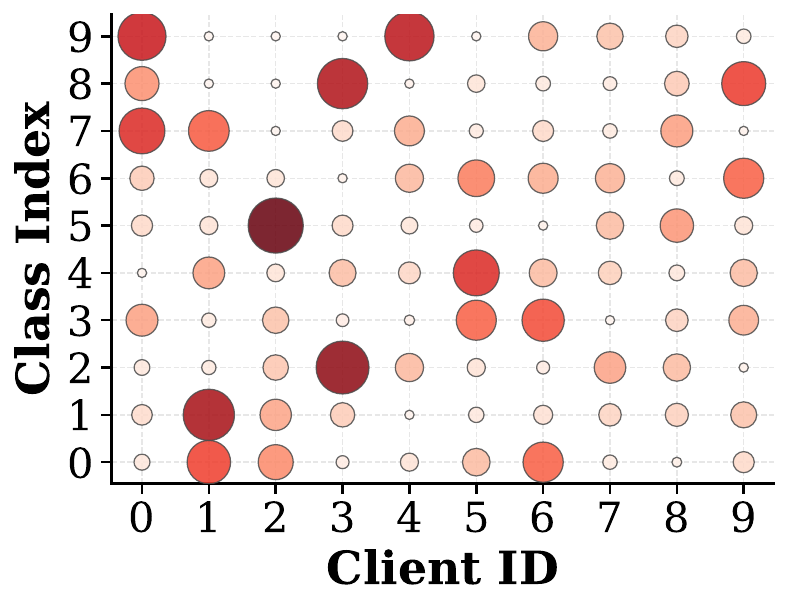}
        \caption{$\text{Dir}(1.0)$}
    \end{subfigure} \\
    \centering
    \begin{subfigure}{0.158\textwidth}
        \centering
        \includegraphics[width=\textwidth]{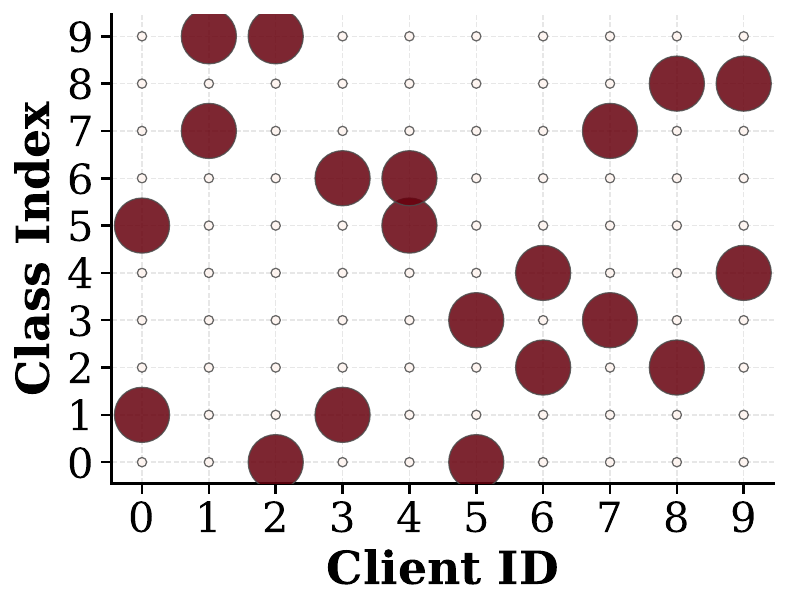}
        \caption{$\text{Path}(2)$}
    \end{subfigure}
    \hfill
    \begin{subfigure}{0.158\textwidth}
        \centering
        \includegraphics[width=\textwidth]{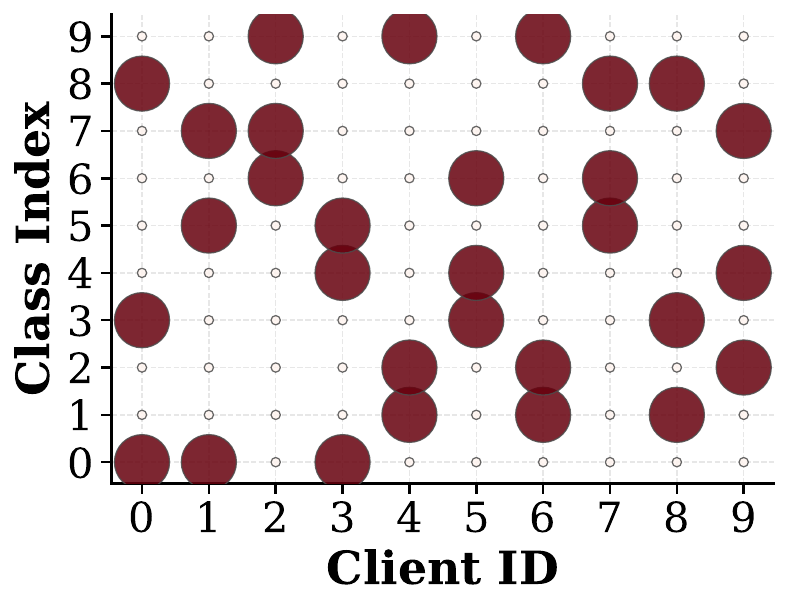}
        \caption{$\text{Path}(3)$}
    \end{subfigure}
    \hfill
    \begin{subfigure}{0.158\textwidth}
        \centering
        \includegraphics[width=\textwidth]{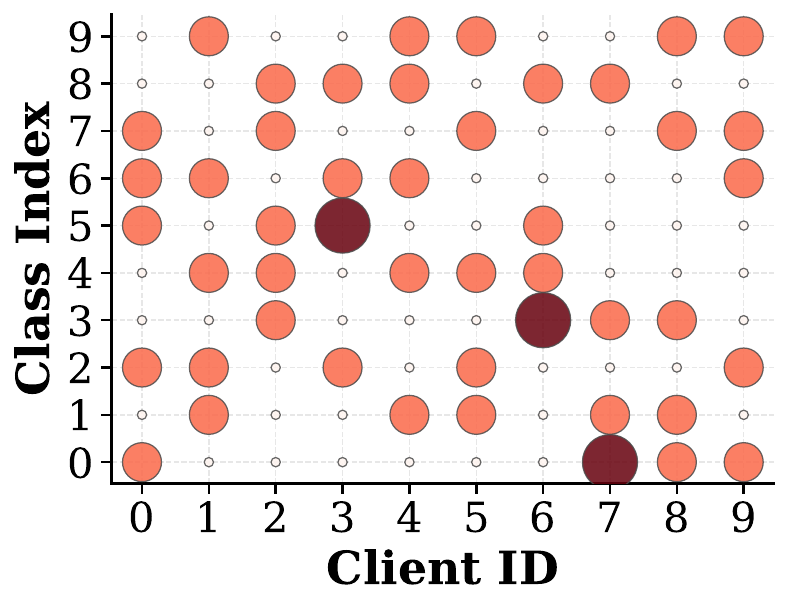}
        \caption{$\text{Path}(5)$}
    \end{subfigure}
    \caption{Data distributions across clients under Dirichlet and Pathological non-\textit{i.i.d.} partitioning protocols.}
    \label{heto}
\end{figure}

\paragraph{Datasets and Models}
We evaluate our approach on three widely-used benchmark datasets: \textsc{CIFAR-10} \cite{cifar}, \textsc{CIFAR-100} \cite{cifar}, and \textsc{SVHN} \cite{svhn}. 
To simulate non-\textit{i.i.d.} data distributions in FL, we follow two widely adopted experimental protocols \cite{DBLP:conf/infocom/WangKNL20,fedavg}, namely the Dirichlet distribution $\text{Dir}(\beta)$ and the Pathological distribution $\text{Path}(r)$, to partition the data.
For the Dirichlet distribution, we consider $\beta \in \{0.1, 0.5, 1.0\}$ in our experiments. 
For the pathological distribution, the parameter $r$ is set to $\{2, 3, 5\}$. 
Smaller values of $\beta$ and $r$ correspond to higher degrees of data heterogeneity. 
Figure \ref{heto} provides a visualization of the data distribution differences under the two partitioning protocols.
In addition, to examine the robustness of \textsc{FLood} across different model architectures, we conduct experiments using two representative neural network models, ResNet-8 \cite{resnet} and MobileNet-V2 \cite{mobilenet}.

\paragraph{FL Implementation}
We consider a FL setup with a total of 100 clients.
To better reflect practical FL scenarios with limited communication resources, only 10\% of clients, corresponding to 10 clients, are randomly selected to participate in each communication round.
The server performs 2000 communication rounds, and each selected client conducts 5 local training epochs per round with a batch size of 50.
All clients adopt the SGD optimizer with a learning rate of 0.01, momentum of 0.9, and weight decay of 0.0005 for local optimization.
In addition, a learning rate decay factor of 0.998 is applied at the end of each round to facilitate stable convergence over training horizons.

\renewcommand{\arraystretch}{1.6}
\begin{table*}[htbp]
    \centering
    \caption{Results of the classification accuracy (mean $\pm$ std) on Dirichlet heterogeneous settings using ResNet-8. 
    All values are expressed as percentages, with the best and second-best results being \textbf{boldfaced} and \underline{underlined}, respectively.}
    \label{main_result_dirichlet}
    \setlength{\tabcolsep}{1.8mm}
    \begin{tabular}{lccccccccc}
    \Xhline{1px}
    \multicolumn{1}{l}{\multirow{2}{*}{\textbf{Baselines}}} & \multicolumn{3}{c}{\textbf{CIFAR-10}} & \multicolumn{3}{c}{\textbf{CIFAR-100}} & \multicolumn{3}{c}{\textbf{SVHN}} \\
    \multicolumn{1}{c}{}                           
    & $Dir(0.1)$   & $Dir(0.5)$ & $Dir(1.0)$ 
    & $Dir(0.1)$   & $Dir(0.5)$ & $Dir(1.0)$
    & $Dir(0.1)$   & $Dir(0.5)$ & $Dir(1.0)$ \\
    \Xhline{0.6px}
    \textsc{FedAvg}
    & $72.79_{(5.95)}$ & $90.18_{(0.25)}$ & $90.82_{(0.11)}$ 
    & $65.51_{(0.24)}$ & $67.26_{(0.13)}$ & $67.59_{(0.11)}$
    & $86.49_{(7.44)}$ & $94.62_{(0.13)}$ & $94.94_{(0.03)}$
    \\
    \Xhline{0.6px}
    
    \textsc{FedProx} 
    & $71.48_{(6.42)}$ & $90.48_{(0.29)}$ & $90.66_{(0.09)}$ 
    & $65.55_{(0.25)}$ & $67.31_{(0.16)}$ & $67.43_{(0.15)}$
    & $84.94_{(7.95)}$ & $94.70_{(0.13)}$ & $94.82_{(0.05)}$\\
   
    \textsc{SCAFFOLD} 
    &$70.61_{(3.55)}$  
    &$90.73_{(0.23)}$  
    &$\underline{91.89}_{(0.20)}$  
    &$67.59_{(0.28)}$  &$70.38_{(0.16)}$  
    &$\underline{70.77}_{(0.11)}$
    &$91.68_{(1.90)}$
    &$95.06_{(0.25)}$  
    &$95.33_{(0.11)}$  \\
    
    \textsc{Moon} 
    & $73.72_{(4.55)}$
    & $90.48_{(0.26)}$
    & $90.78_{(0.11)}$
    & $65.90_{(0.23)}$
    & $67.65_{(0.14)}$
    & $67.00_{(0.14)}$
    & $85.83_{(7.25)}$
    & $94.79_{(0.13)}$
    & $94.89_{(0.04)}$
    \\  
    
    \textsc{FedDecorr}
    & $70.53_{(6.69)}$
    & $90.34_{(0.29)}$
    & $90.57_{(0.09)}$
    & $64.21_{(0.26)}$
    & $66.11_{(0.14)}$
    & $65.70_{(0.16)}$
    & $85.04_{(8.78)}$
    & $94.60_{(0.13)}$
    & $94.65_{(0.06)}$ \\  
    
    \textsc{FedNP} 
    &$66.60_{(8.28)}$  &$89.57_{(0.48)}$  
    &$90.41_{(0.17)}$  
    &$64.78_{(0.31)}$  
    &$67.18_{(0.14)}$  
    &$67.39_{(0.14)}$ 
    &$82.44_{(10.49)}$  
    &$94.17_{(0.30)}$  
    &$94.56_{(0.10)}$  \\  
    

    \Xhline{0.6px}
    
    \textsc{FedAvgM} 
    & $70.29_{(7.52)}$ & $90.51_{(0.26)}$ & $90.95_{(0.11)}$
    & $65.53_{(0.22)}$
    & $67.59_{(0.15)}$
    & $67.22_{(0.11)}$ 
    & $85.42_{(7.57)}$
    & $94.85_{(0.13)}$
    & $94.93_{(0.05)}$ \\
    
    \textsc{FedNova}
    & $71.16_{(6.56)}$
    & $90.26_{(0.30)}$
    & $91.09_{(0.13)}$
    & $66.09_{(0.23)}$
    & $67.60_{(0.10)}$
    & $67.48_{(0.13)}$
    & $84.93_{(5.66)}$
    & $94.79_{(0.14)}$
    & $94.89_{(0.05)}$\\
    
    \textsc{FedDisco} 
    & $\underline{75.75}_{(4.66)}$ & $90.32_{(0.34)}$ & $90.73_{(0.14)}$ & $65.75_{(0.23)}$ & $67.35_{(0.13)}$ & $67.85_{(0.14)}$
    & $89.25_{(3.82)}$ & $94.72_{(0.11)}$ & $95.03_{(0.06)}$
    \\
    
    \textsc{FedMR} 
    &$36.03_{(6.19)}$  &$\underline{90.74}_{(0.24)}$  
    &$91.69_{(0.13)}$  
    &$\underline{67.68}_{(0.20)}$  
    &$\underline{70.56}_{(0.10)}$  
    &$69.88_{(0.12)}$  
    &$\underline{92.91}_{(1.37)}$
    &$\underline{95.60}_{(0.05)}$  
    &$\underline{95.76}_{(0.04)}$  \\
    \Xhline{0.6px}
    
    \textsc{FLood (Ours)} 
    & $\textbf{84.43}_{(1.66)}$ & $\textbf{92.92}_{(0.09)}$ & $\textbf{93.28}_{(0.08)}$ 
    & $\textbf{70.98}_{(0.13)}$ & $\textbf{71.30}_{(0.17)}$ & $\textbf{71.44}_{(0.12)}$ & $\textbf{94.50}_{(0.34)}$ & $\textbf{95.67}_{(0.05)}$ & $\textbf{95.85}_{(0.04)}$\\
    \Xhline{1px}
    \end{tabular}
\end{table*}

\renewcommand{\arraystretch}{1.6}
\begin{table*}[htbp]
    \centering
    \caption{Results of the classification accuracy (mean $\pm$ std) on Pathological heterogeneous settings using ResNet-8. 
    All values are expressed as percentages, with the best and second-best results being \textbf{boldfaced} and \underline{underlined}, respectively.}
    \label{main_result_pathological}
    \setlength{\tabcolsep}{1.8mm}
    \begin{tabular}{lccccccccc}
    \Xhline{1px}
    \multicolumn{1}{l}{\multirow{2}{*}{\textbf{Baselines}}} & \multicolumn{3}{c}{\textbf{CIFAR-10}} & \multicolumn{3}{c}{\textbf{CIFAR-100}} & \multicolumn{3}{c}{\textbf{SVHN}} \\
    \multicolumn{1}{c}{}                           
    & $\text{Path}(2)$   & $\text{Path}(3)$ & $\text{Path}(5)$   & $\text{Path}(2)$   & $\text{Path}(3)$ & $\text{Path}(5)$  & $\text{Path}(2)$   & $\text{Path}(3)$ & $\text{Path}(5)$ \\ \Xhline{0.6px}
    \textsc{FedAvg} 
    & $43.53_{(4.53)}$ & $71.33_{(5.55)}$ & $89.72_{(0.51)}$
    & $8.64_{(2.20)}$ & $26.22_{(3.45)}$ & $53.62_{(1.66)}$ & $80.00_{(7.59)}$ & $91.35_{(3.67)}$ & $94.76_{(0.23)}$\\
    \Xhline{0.6px}
    
    \textsc{FedProx} 
    & $44.68_{(4.50)}$ & $72.55_{(5.43)}$ & $89.80_{(0.57)}$ & $9.24_{(2.01)}$ & $27.58_{(3.08)}$ &$53.44_{(1.67)}$
    & $79.39_{(7.84)}$ & $91.75_{(3.48)}$ & $94.73_{(0.24)}$\\
   
    \textsc{SCAFFOLD}
    &$50.16_{(5.17)}$
    &$73.67_{(4.51)}$  
    &$\underline{90.27}_{(0.57)}$  
    &$9.04_{(2.20)}$ 
    &$29.49_{(2.78)}$ &$54.56_{(1.35)}$
    &$79.50_{(7.08)}$ 
    &$91.39_{(4.23)}$ &$95.26_{(0.10)}$\\
    
    \textsc{Moon} &$45.05_{(5.07)}$ &$72.71_{(5.40)}$ &$89.83_{(0.46)}$ 
    &$8.49_{(2.37)}$ 
    &$26.86_{(3.17)}$ 
    &$54.09_{(1.59)}$ &$78.39_{(7.64)}$ 
    &$91.41_{(3.46)}$ 
    &$94.71_{(0.30)}$\\  
    
    \textsc{FedDecorr}
    & $\underline{61.21}_{(5.15)}$ & $\underline{78.42}_{(3.94)}$ & $89.73_{(0.42)}$ 
    & $\underline{12.02}_{(2.69)}$ & $\underline{30.11}_{(3.28)}$ &$\underline{55.32}_{(0.87)}$
    & $83.13_{(6.42)}$ & $92.50_{(2.34)}$ & $94.53_{(0.37)}$
    \\
    
    \textsc{FedNP}
    & $43.74_{(5.54)}$ 
    & $70.16_{(6.09)}$ 
    & $88.66_{(0.99)}$ 
    & $8.50_{(2.90)}$ 
    & $26.69_{(3.14)}$ 
    & $51.42_{(1.83)}$
    & $78.30_{(8.86)}$ 
    & $90.33_{(4.58)}$ 
    & $94.18_{(0.36)}$
    \\
    
    \Xhline{0.6px}
    
    \textsc{FedAvgM}
    & $42.29_{(4.07)}$ & $71.60_{(5.71)}$ & $90.08_{(0.59)}$ & $7.47_{(1.95)}$ & $24.63_{(3.31)}$ & $53.64_{(1.68)}$ & $77.82_{(8.07)}$ & $90.97_{(3.71)}$ & $94.72_{(0.45)}$ \\
    
    \textsc{FedNova} & $44.70_{(4.26)}$ 
    & $72.29_{(5.60)}$ & $89.73_{(0.59)}$ 
    & $9.00_{(2.19)}$ & $25.89_{(3.30)}$
    & $52.94_{(1.63)}$ & $76.93_{(7.76)}$ & $90.27_{(4.14)}$ & $94.86_{(0.20)}$
    \\
    
    \textsc{FedDisco} 
    & $43.05_{(4.18)}$ & $72.62_{(5.14)}$ & $89.40_{(0.47)}$
    & $8.62_{(2.19)}$ & $28.14_{(3.30)}$ & $54.21_{(1.61)}$ 
    & $79.64_{(7.45)}$ & $91.65_{(3.57)}$ & $94.75_{(0.17)}$
    \\
    
    \textsc{FedMR} 
    &$40.08_{(3.79)}$ 
    &$76.72_{(1.32)}$ 
    &$88.59_{(0.53)}$
    &$2.88_{(0.52)}$ 
    &$4.21_{(0.71)}$ 
    &$32.51_{(1.71)}$
    &$\textbf{87.99}_{(5.08)}$
    &$\textbf{94.43}_{(0.89)}$ 
    &$\textbf{95.89}_{(0.05)}$\\
    \Xhline{0.6px}
    
    \textsc{FLood (Ours)} 
    & $\textbf{62.37}_{(4.30)}$ & $\textbf{86.16}_{(2.96)}$ & $\textbf{93.34}_{(0.09)}$ 
    & $\textbf{15.31}_{(2.43)}$ & $\textbf{35.52}_{(2.39)}$ & $\textbf{61.05}_{(0.69)}$ & $\underline{86.55}_{(6.48)}$ & $\underline{93.67}_{(2.73)}$ & $\underline{95.78}_{(0.04)}$\\
    \Xhline{1px}
    \end{tabular}
    \label{main_result}
\end{table*}

\subsection{Comparison with state-of-the-art Baselines (RQ1)}
\label{5.2}
Tables \ref{main_result_dirichlet} and \ref{main_result_pathological} present a comprehensive performance comparison between \textsc{FLood} and representative FL baselines under two heterogeneity protocols across three benchmark datasets. 
To ensure fair and stable evaluation, we report the average classification accuracy and standard deviation computed over the final 50 communication rounds.
From Tables \ref{main_result_dirichlet} and \ref{main_result_pathological}, it is evident that \textsc{FLood} consistently achieves the best or near-best performance across all heterogeneous settings and datasets. 
Under the Dirichlet distribution, \textsc{FLood} demonstrates clear and consistent improvements over both client-side correction methods and server-side aggregation methods. 
For example, on the \textsc{CIFAR-10} dataset with $\text{Dir}(0.1)$, \textsc{FLood} outperforms the second-best baseline by 8.68\% in accuracy, highlighting its strong capability to mitigate severe data heterogeneity. 
Similar performance margins are observed on \textsc{CIFAR-100} and \textsc{SVHN}, indicating the robustness of \textsc{FLood} across datasets with varying complexity.
Notably, the performance gains of \textsc{FLood} are more pronounced under highly heterogeneous settings, such as $\text{Dir}(0.1)$ and $\text{Path}(2)$, compared to milder scenarios like $\text{Dir}(1.0)$ or $\text{Path}(5)$.
This trend suggests that \textsc{FLood} is particularly effective when data heterogeneity poses a significant challenge to federated optimization.
A plausible explanation is that, in scenarios with low data heterogeneity, the discrepancy between genuine-ID and pseudo-OOD samples becomes less prominent, thereby reducing the potential benefit of OOD-aware reweighting. 
In contrast, under highly heterogeneous conditions, \textsc{FLood} can more accurately identify distributionally misaligned samples and unreliable client updates, leading to substantially larger performance improvements.
We also observe that \textsc{FedMR} achieves competitive or slightly superior performance in a few settings. 
However, the performance gap with \textsc{FLood} in these cases is relatively small. 
In contrast, across the majority of evaluated scenarios, particularly under severe data heterogeneity, \textsc{FLood} consistently and noticeably outperforms \textsc{FedMR}. 

\begin{figure}[htbp]
    \centering
    \begin{subfigure}{0.48\textwidth}
        \centering
        \includegraphics[width=\textwidth]{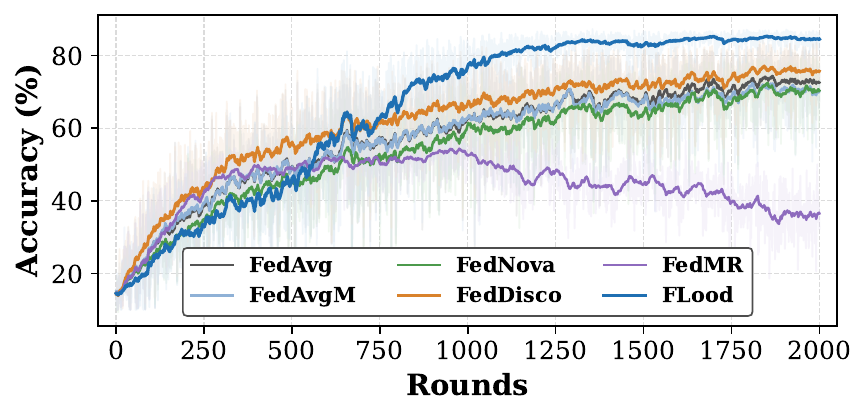}
        \caption{vs. server-side methods.}
    \end{subfigure} \\
    \begin{subfigure}{0.48\textwidth}
        \centering
        \includegraphics[width=\textwidth]{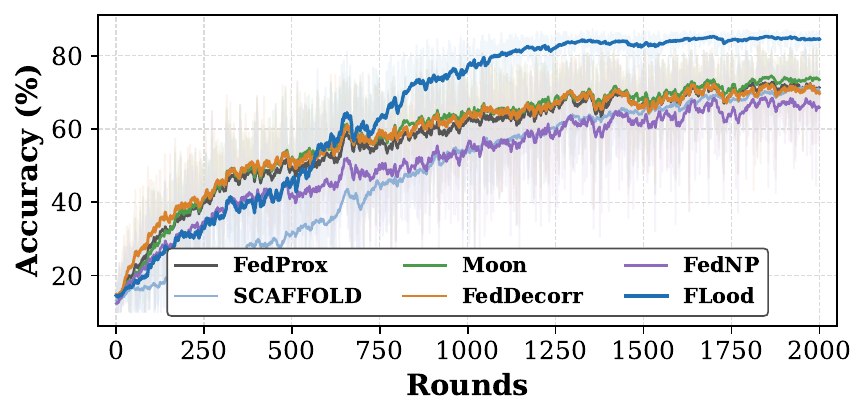}
        \caption{vs. client-side methods.}
    \end{subfigure}
    \caption{Accuracy curves of various FL methods on \textsc{CIFAR-10} dataset under the $\text{Dir}(0.1)$ scenario. 
    On the top, \textsc{FLood} is compared with \textsc{FedAvg} and server-side global adjustment methods, while on the bottom, it is compared with client-side local training correction methods.}
    \label{accuracy_trend}
\end{figure}

We further illustrate the training dynamics of different FL methods in Figure \ref{accuracy_trend}. 
As shown in both comparisons, the accuracy curves reveal a characteristic learning trajectory for \textsc{FLood}.
In the early rounds, \textsc{FLood} initially lags behind most baselines. 
This behavior is expected, as the global model at this stage has limited discriminative capability and its OOD detection signal is still relatively immature. 
Consequently, the distinction between genuine-ID and pseudo-OOD samples is less reliable, which may lead to suboptimal weight assignments during local updates.
As training progresses into the mid stage, the model becomes better aligned with the underlying data distribution, allowing \textsc{FLood} to more accurately identify pseudo-OOD samples and to adjust their weights accordingly. 
This results in performance that gradually matches, and eventually surpasses that of existing methods.
In the later training rounds, \textsc{FLood} demonstrates a clear and consistent advantage.
The increasingly reliable OOD confidence estimates enable more targeted emphasis on challenging samples.
The overall trend confirms that \textsc{FLood} benefits from its progressive learning mechanism: once the OOD signals stabilize, the method leverages them effectively to achieve superior convergence behavior and final performance.

\begin{figure}[htbp]
    \centering
    \begin{subfigure}{0.24\textwidth}
        \centering
        \includegraphics[width=\textwidth]{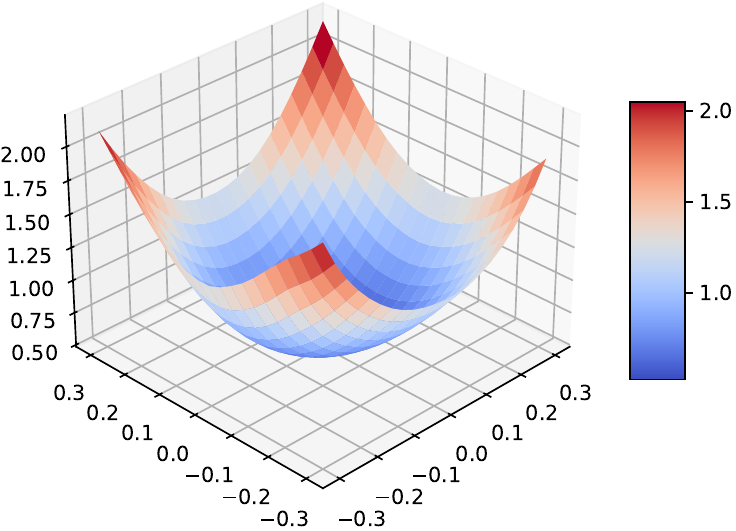}
        \caption{\textsc{FedAvg}-$\text{Dir}(0.1)$}
        \label{fig:loss-1}
    \end{subfigure}
    \hfill
    \begin{subfigure}{0.24\textwidth}
        \centering
        \includegraphics[width=\textwidth]{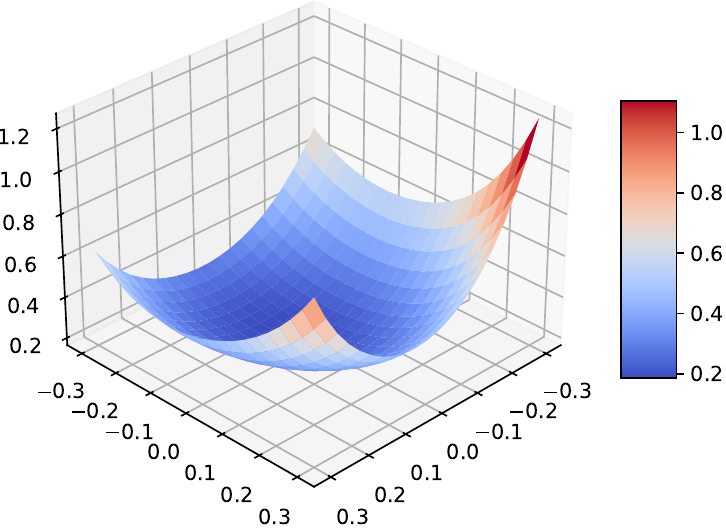}
        \caption{\textsc{FLood}-$\text{Dir}(0.1)$}
        \label{fig:loss-2}
    \end{subfigure}
    \hfill \\
    \begin{subfigure}{0.24\textwidth}
        \centering
        \includegraphics[width=\textwidth]{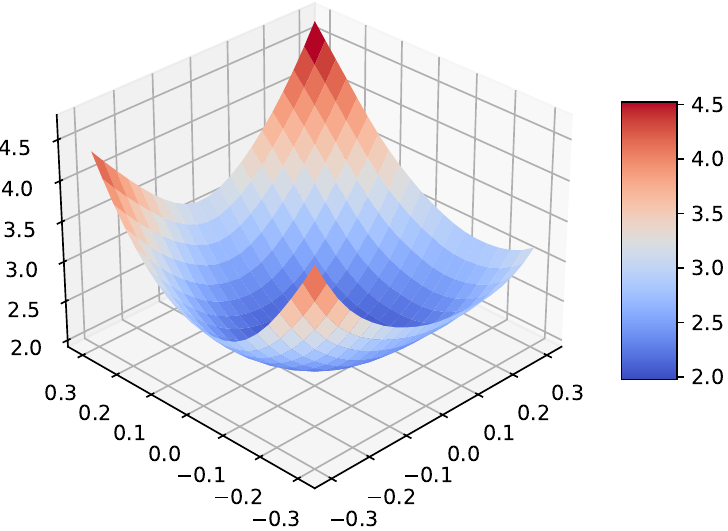}
        \caption{\textsc{FedAvg}-$\text{Path}(2)$}
        \label{fig:loss-3}
    \end{subfigure}
    \hfill
    \begin{subfigure}{0.24\textwidth}
        \centering
        \includegraphics[width=\textwidth]{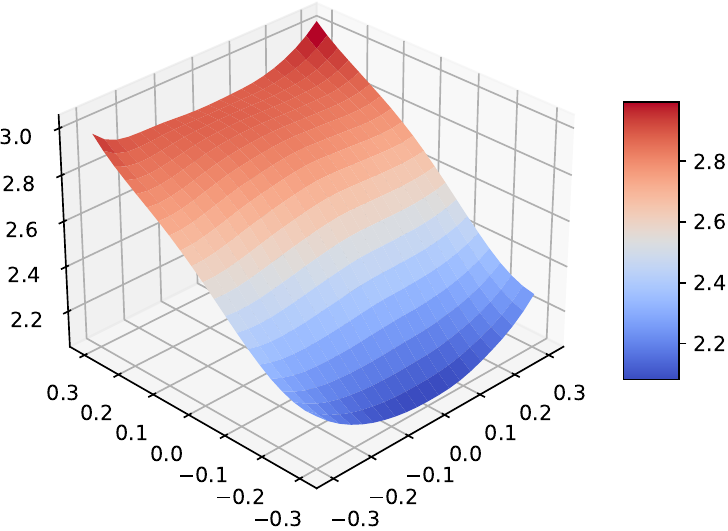}
        \caption{\textsc{FLood}-$\text{Path}(2)$}
        \label{fig:loss-4}
    \end{subfigure}
    \caption{Loss landscapes of \textsc{FedAvg} and \textsc{FLood}.}
    \label{loss-landscapes}
\end{figure}

Figure \ref{loss-landscapes} visualizes the loss landscapes of models trained with \textsc{FedAvg} and \textsc{FLood} under different levels of data heterogeneity.
Across all settings, \textsc{FedAvg} consistently converges to sharp minima with steep curvature along multiple directions, suggesting that the associated Hessian spectrum contains relatively large eigenvalues.
In contrast, \textsc{FLood} produces substantially flatter minima, characterized by broader valley regions and noticeably reduced curvature.
From the perspective of \cite{flat}, flat minima indicate that small perturbations in model parameters result in only marginal increases in loss, which is widely recognized as a strong indicator of improved generalization.
These landscape visualizations therefore provide clear geometric evidence that \textsc{FLood} attains more stable, robust, and better generalizable solutions than \textsc{FedAvg} under non-\textit{i.i.d.} data distributions.

\begin{figure}[htbp]
    \centering
    \begin{subfigure}{0.24\textwidth}
        \centering
        \includegraphics[width=\textwidth]{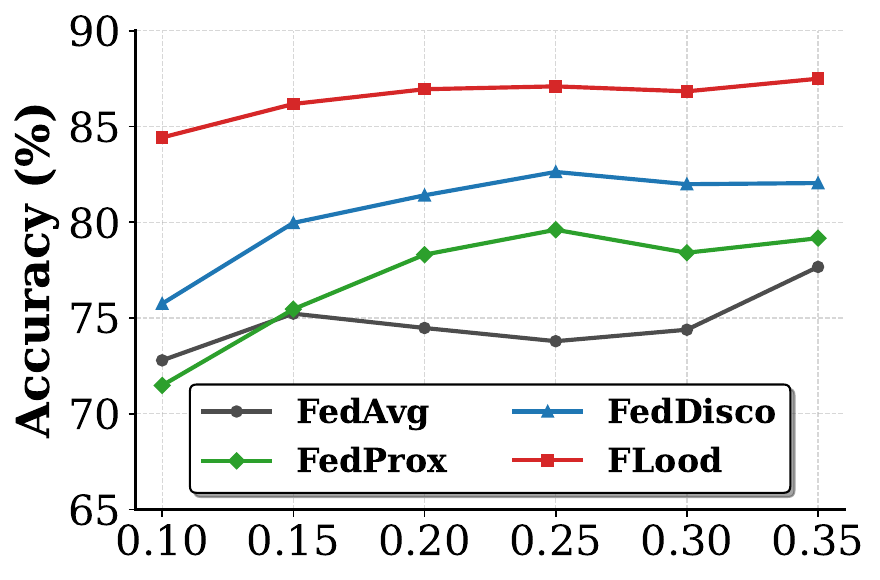}
        \caption{Fraction.}
        \label{fig:fraction}
    \end{subfigure}
    \hfill
    \begin{subfigure}{0.24\textwidth}
        \centering
        \includegraphics[width=\textwidth]{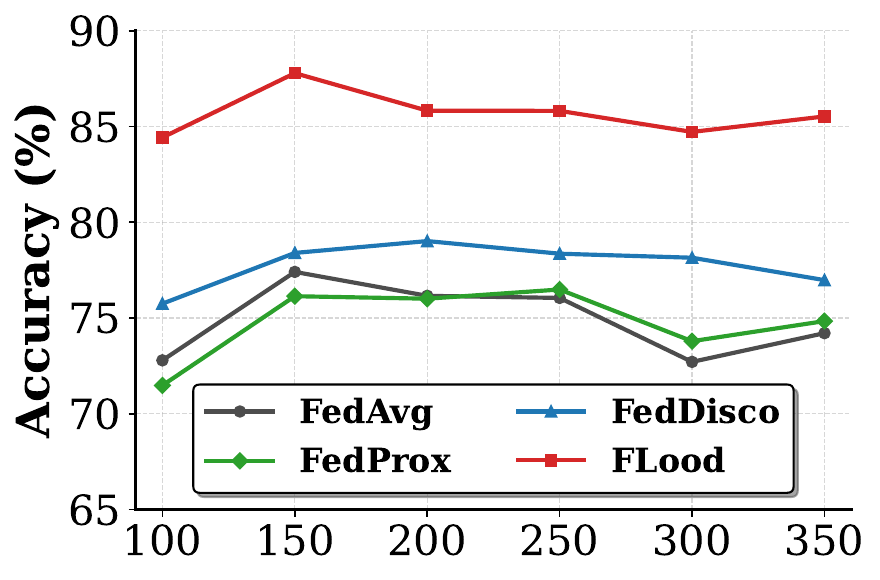}\caption{Number of clients.}
        \label{fig:client_number}
    \end{subfigure}
    \caption{Impact of participation fraction and client number.}
    \label{scalability}
\end{figure}

\subsection{Scalability Analysis (RQ2)}
Figure \ref{scalability} evaluates the scalability of \textsc{FLood} with respect to two key system-level parameters, namely the client participation fraction and the total number of clients.
In Figure \ref{fig:fraction}, we vary the fraction of participating clients per communication round while keeping all other settings fixed.
The results show that \textsc{FLood} consistently outperforms \textsc{FedAvg}, \textsc{FedProx}, and \textsc{FedDisco} across all participation fractions, demonstrating strong robustness to partial client availability.
Moreover, the performance of \textsc{FLood} remains stable as the participation rate increases, indicating that it can effectively exploit additional client updates without introducing aggregation instability or performance degradation.
Figure \ref{fig:client_number} further examines scalability by increasing the total number of clients in the system.
As the client population grows, \textsc{FLood} continues to achieve robust and stable accuracy, while the baseline methods exhibit larger performance fluctuations. 
This behavior suggests that the OOD-aware aggregation mechanism in \textsc{FLood} is effective at selectively amplifying reliable client updates, thereby maintaining stable learning dynamics under increased system scale and client diversity.
Overall, these results demonstrate that \textsc{FLood} scales favorably with respect to both the participation rate and total client number of clients, highlighting its robustness to system-level variability and its practicality for large-scale FL deployments.

\subsection{Compatibility with client-side FL Methods (RQ3)}
By directly modifying the local supervised loss, \textsc{FLood} can be seamlessly incorporated as an orthogonal plug-in module to enhance existing client-side local training correction methods. 
Table \ref{Orthogonality} presents the results of the orthogonal compatibility experiments under different pathological heterogeneity settings. 
For each baseline method, the table reports the absolute classification accuracy achieved after integrating \textsc{FLood}, along with the corresponding performance gains over the original baseline under each heterogeneity level. 
The results clearly demonstrate that \textsc{FLood} consistently improves the performance of all evaluated client-side methods, confirming its strong compatibility and complementary nature. 
Notably, when combined with \textsc{FedProx}, \textsc{FLood} yields a substantial accuracy improvement of up to 17.77\% in $\text{Path}(2)$, highlighting the effectiveness of OOD-aware sample reweighting in correcting biased local updates. 
In addition, \textsc{FLood} achieves particularly strong performance when integrated with \textsc{FedDecorr}, indicating that it can further amplify the benefits of advanced local training correction strategies across a wide range of heterogeneity settings.

\renewcommand{\arraystretch}{1.6}
\begin{table}[htbp]
    \centering
    \setlength{\tabcolsep}{2.2mm}
    \begin{tabular}{ccccccc}
    \Xhline{1px}
    \textbf{Methods} & Path(2) & $\uparrow$ & Path(3) & $\uparrow$ & Path(5) & $\uparrow$ \\
    \hline
    \textsc{FedProx} 
        & 62.45 & \textbf{17.77} 
        & 84.80 & \textbf{12.25} 
        & 92.83 & \textbf{3.03} \\

    \textsc{SCAFFOLD} 
        & 58.20 & \textbf{8.04} 
        & 81.52 & \textbf{7.85} 
        & 92.98 & \textbf{2.71} \\

    \textsc{FedDecorr} 
        & 69.31 & \textbf{8.10} 
        & 87.57 & \textbf{9.15} 
        & 92.99 & \textbf{3.26} \\

    \textsc{FedNP} 
        & 59.35 & \textbf{15.61} 
        & 86.09 & \textbf{15.93} 
        & 92.19 & \textbf{3.53} \\
    \Xhline{1px}
    \end{tabular}
    \caption{Orthogonal experiments comparing \textsc{FLood} with client-side local training correction methods on the \textsc{CIFAR-10} dataset under pathological heterogeneity.}
    \label{Orthogonality}
\end{table}

\subsection{Ablation Study (RQ4)}

\begin{figure}[htbp]
    \centering
    \begin{subfigure}{0.24\textwidth}
        \centering
        \includegraphics[width=\textwidth]{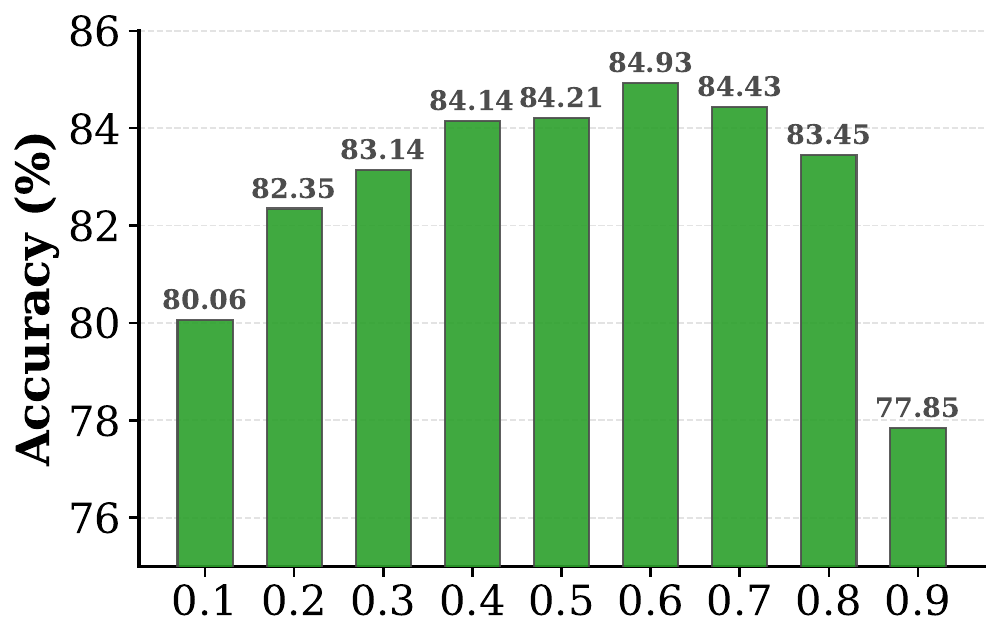}
        \caption{OOD $q^{th}$ threshold.}
        \label{fig:q}
    \end{subfigure}
    \hfill
    \begin{subfigure}{0.24\textwidth}
        \centering
        \includegraphics[width=\textwidth]{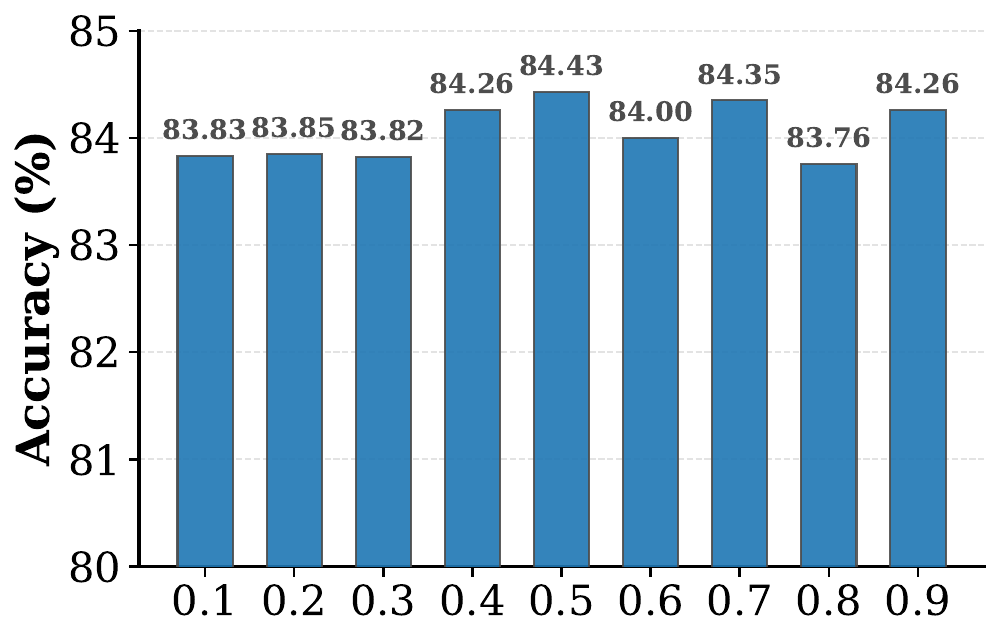}       \caption{Aggregation ratio $\alpha$.}
        \label{fig:alpha}
    \end{subfigure}\\
    \vspace{0.1in}
    \begin{subfigure}{0.238\textwidth}
        \centering
        \includegraphics[width=\textwidth]{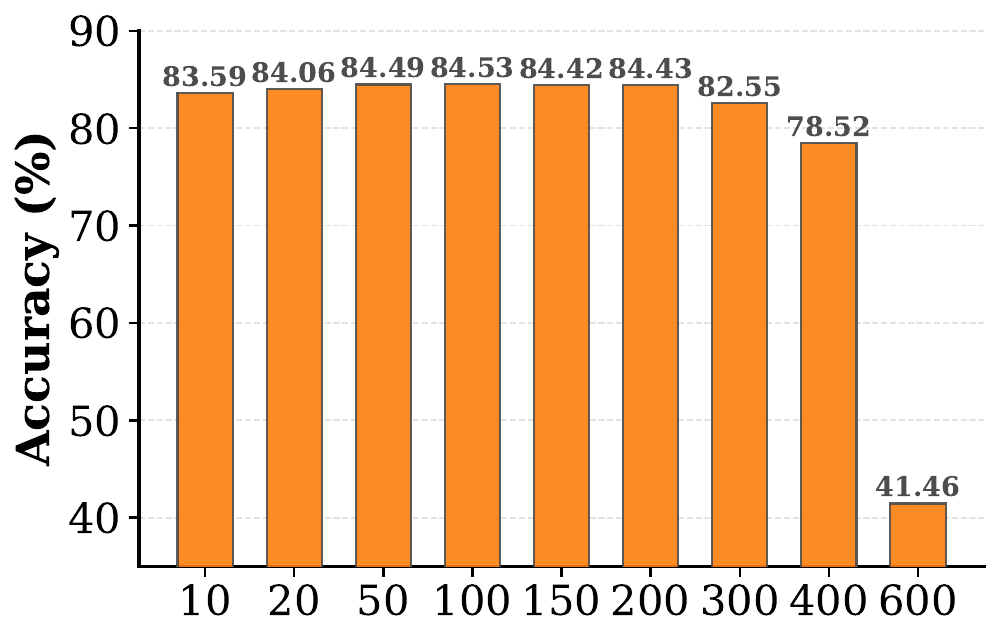}
        \caption{Amplification factor $a$.}
        \label{fig:a}
    \end{subfigure}
    \hfill
    \begin{subfigure}{0.238\textwidth}
        \centering
        \includegraphics[width=\textwidth]{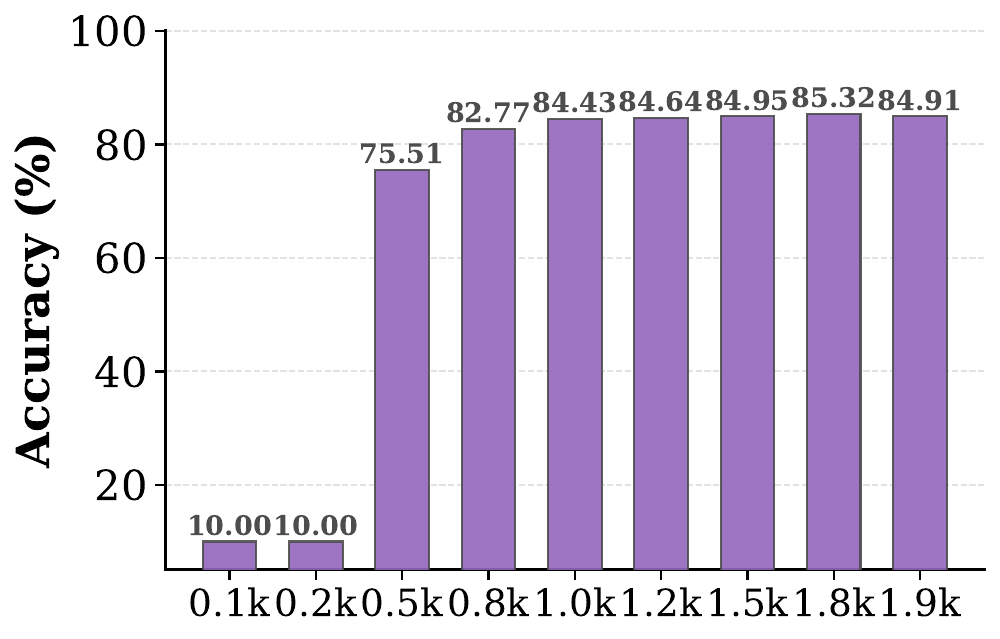}       
        \caption{Stabilization round $T$.}
        \label{fig:t}
    \end{subfigure}
    \caption{Effects of four key hyperparameters.}
\end{figure}

\paragraph{Effect of hyperparameters}
We analyze the sensitivity of \textsc{FLood} to four critical hyperparameters, namely $q$, $a$, $T$, and $\alpha$.
Figure \ref{fig:q} illustrates the impact of the local OOD threshold $q$ on the performance of \textsc{FLood}. 
As shown in the figure, \textsc{FLood} achieves stable and strong performance when $q$ lies in the range $[0.2, 0.8]$, indicating robustness to moderate threshold variations. 
When $q$ is too small, such as $q = 0.1$, performance degrades sharply because an excessive number of samples are classified as pseudo-OOD, which overly suppresses genuine-ID samples and hinders effective supervised learning.
In contrast, overly large thresholds also lead to mild performance degradation, as insufficient emphasis is placed on pseudo-OOD samples.

To examine the effect of server-side aggregation, Figure \ref{fig:alpha} presents a grid search over the aggregation ratio $\alpha$.
The results indicate that \textsc{FLood} achieves optimal performance when $\alpha = 0.5$, while deviations from this value lead to mildly degradation.
When $\alpha is$ too small, the influence of OOD confidence is limited and aggregation is largely dominated by data volume. 
In contrast, excessively large values of $\alpha$ may overemphasize a subset of clients, resulting in biased and suboptimal aggregation.

Figure \ref{fig:a} explores how the amplification factor $a$ influences the behavior of \textsc{FLood}. 
The results show that performance improves as $a$ increases from small values, indicating that amplifying pseudo-OOD samples is beneficial. 
However, once $a$ exceeds a certain threshold, accuracy begins to deteriorate. 
This suggests that overly aggressive amplification can dominate the learning dynamics, causing the model to overfocus on pseudo-OOD samples and lose balance in supervised training.

We investigate the temporal aspect of weight scheduling in Figure \ref{fig:t}. 
When $T$ is small, strong reweighting is introduced prematurely, leading to unstable training and inferior performance. 
Increasing $T$ gradually improves performance, reflecting the importance of postponing OOD-based reweighting until the model becomes sufficiently reliable.
Beyond a certain point, further increases in $T$ yield limited gains, suggesting that the benefits of delayed amplification eventually saturate.

Taken together, these results demonstrate that the effectiveness of \textsc{FLood} depends on the coordinated tuning of its core hyperparameters, which jointly balance training stability and robustness under heterogeneous data distributions.


\renewcommand{\arraystretch}{1.6}
\begin{table}[h]
\setlength{\tabcolsep}{2.2mm}
\caption{Impact of two components across two models.}
\label{ablation}
\begin{tabular}{ccccc}
\Xhline{1px}
& \textsc{FedAvg} & \textsc{+DAC} & \textsc{+ASW}  & \textsc{FLood}  \\ \hline   
ResNet & $72.79_{(5.95)}$   & $74.46_{(4.62)}$    & $82.71_{(2.10)}$ & $\textbf{84.43}_{(1.66)}$     \\
MobileNet & $55.01_{(7.36)}$   & $56.99_{(5.95)}$    & $63.40_{(6.80)}$ & $\textbf{66.57}_{(3.92)}$ \\
\Xhline{1px}
\end{tabular}
\end{table}

\paragraph{Impact of each component}
Table \ref{ablation} reports the ablation results of \textsc{FLood} by comparing its decoupled components across two model architectures, where ASW denotes the adaptive sample weighting and DAC denotes the dynamic aggregation correction. 
The results demonstrate that both components contribute positively to performance improvements over the \textsc{FedAvg} baseline. 
In particular, ASW alone yields substantial gains, indicating that emphasizing pseudo-OOD samples during local training is effective in mitigating biased local updates. 
Similarly, incorporating DAC improves performance by refining server-side aggregation based on client reliability.
More importantly, combining ASW and DAC consistently achieves the best performance across both ResNet and MobileNet architectures, significantly outperforming either component in isolation. 

\begin{figure}[htbp]
    \centering
    \begin{subfigure}{0.24\textwidth}
        \centering
        \includegraphics[width=\textwidth]{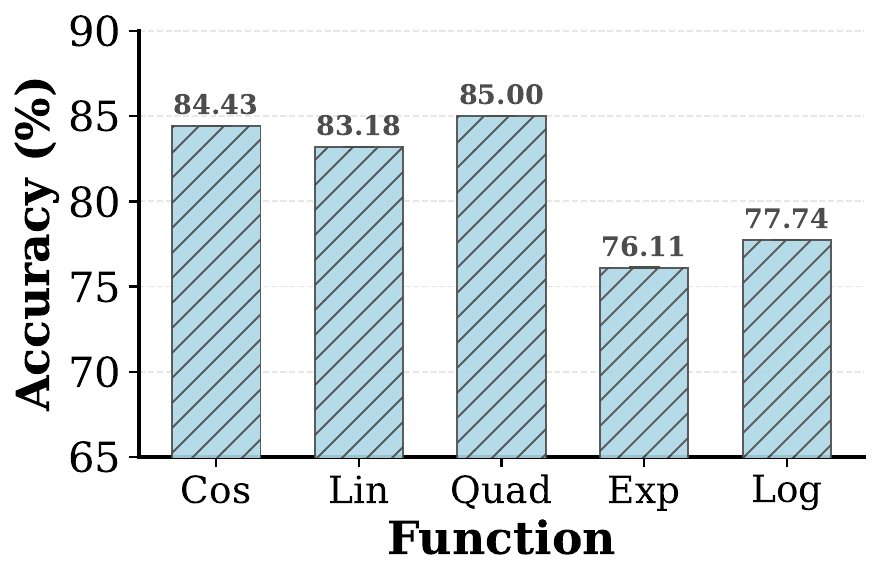}
        \caption{Scheduling functions.}
        \label{5-function}
    \end{subfigure}
    \hfill
    \begin{subfigure}{0.24\textwidth}
        \centering
        \includegraphics[width=\textwidth]{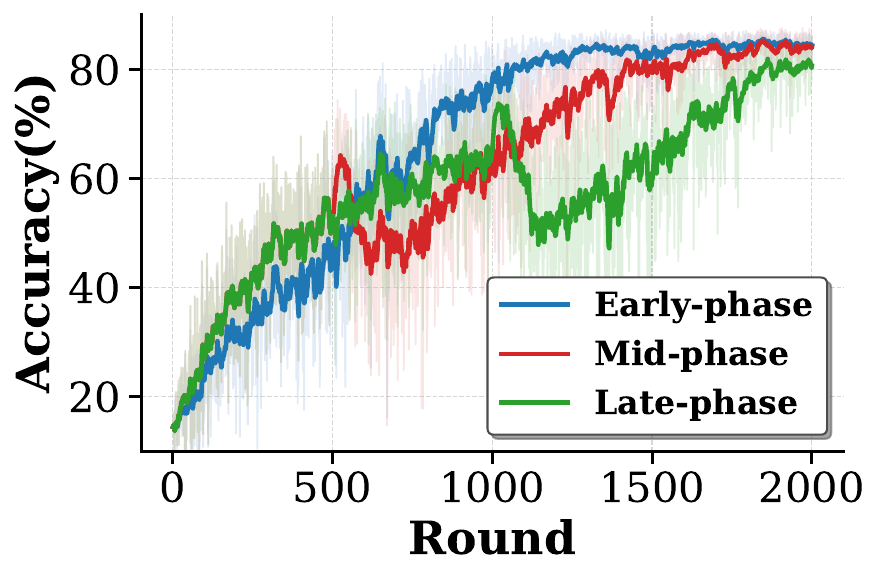}
        \caption{Increasing phases.}
        \label{3-phase}
    \end{subfigure}
    \caption{Impact of Scheduling Strategy on 
    \textsc{FLood}.}
\end{figure}

\paragraph{Impact of weight scheduling strategy}
Figure \ref{5-function} compares different scheduling functions and shows that smooth, progressive schemes such as cosine and quadratic consistently outperform more aggressive alternatives like exponential scheduling, indicating that gradual weight amplification is essential for stable optimization.
Figure \ref{3-phase} further analyzes different growth phase designs and reveals that activating weight growth at an appropriate stage leads to better convergence, whereas starting too early suffers from unreliable OOD estimates and starting too late limits the benefit of emphasizing challenging samples.
Overall, these results highlight that both the functional form and the timing of weight growth play a critical role in effectively leveraging OOD signals within \textsc{FLood}.

\renewcommand{\arraystretch}{1.6}
\begin{table}[h]
\setlength{\tabcolsep}{0.9mm}
\caption{Impact of different OOD methods on \textsc{FLood}.}
\label{ood_methods}
\begin{tabular}{cccccccc}
\Xhline{1px}
& \textsc{MSP} & \textsc{MaxLogit} & \textsc{Energy}  & \textsc{GEN}  & \textsc{ASH-S} & \textsc{ASH-B} & \textsc{ASH-P} \\ \hline   
$\text{Dir}(0.1)$ & $84.45$   & $84.81$    & $84.43$ & $85.86$     & $\underline{10.00}$    &$74.79$ &$84.50$  \\
$\text{Dir}(0.5)$ & $93.44$   & $93.40$    & $92.92$ & $93.11$     & $93.13$    &$92.63$ &$93.02$  \\
$\text{Dir}(1.0)$ & $93.37$   & $93.39$    & $93.28$ & $93.42$     & $93.48$    &$93.54$ &$93.51$  \\
$\text{Path}(2)$ & $61.55$   & $56.50$    & $62.37$ & $55.43$     & $\underline{10.00}$    &$56.97$ &$61.05$  \\
$\text{Path}(3)$ & $86.35$   & $85.96$    & $86.16$ & $84.42$     & $85.01$    &$84.29$ &$85.12$  \\
$\text{Path}(5)$ & $92.94$   & $93.00$    & $93.34$ & $93.01$     & $93.15$    &$92.56$ &$93.07$  \\
   \Xhline{1px}
\end{tabular}
\end{table}

\paragraph{Influence of OOD methods}
We further investigate the influence of different OOD detection methods on the performance of \textsc{FLood}, with the results summarized in Table \ref{ood_methods}. 
Specifically, we evaluate a diverse set of representative OOD scoring functions, including \textsc{MSP} \cite{MSP}, \textsc{MaxLogit} \cite{MaxLogit}, \textsc{Energy} \cite{Energy}, \textsc{GEN} \cite{GEN}, and the \textsc{ASH} variants (\textsc{ASH-S}, \textsc{ASH-B}, and \textsc{ASH-P}) \cite{ASH}.
Overall, most OOD detection methods lead to clear performance improvements over the \textsc{FedAvg} baseline across both heterogeneity settings, demonstrating that \textsc{FLood} is largely agnostic to the specific choice of OOD scoring function in practical scenarios.
Notably, certain OOD methods such as \textsc{ASH-S}, perform poorly under severe heterogeneity (e.g., Dir$(0.1)$ and Path$(2)$), leading to performance collapse. 
This suggests that overly aggressive or unstable OOD scoring may introduce noise into the reweighting process. 
In contrast, stable scoring functions such as \textsc{Energy}, \textsc{MSP}, and \textsc{MaxLogit} yield consistently strong and reliable results across all settings.
\section{Conclusion}
\label{sec:conclusion}
The critical challenge of data heterogeneity in federated learning, arising from the inherently non-\textit{i.i.d.} nature of client data, remains a significant barrier to the development of robust and reliable global models.
This work addresses this challenge by introducing \textsc{FLood}, a novel FL framework that leverages out-of-distribution detection to dynamically mitigate the adverse effects of data heterogeneity.
\textsc{FLood} incorporates a dual-weighting strategy that adaptively adjusts sample weights on the client side and aggregation weights on the server side, thereby effectively countering the inconsistencies introduced by non-\textit{i.i.d.} data.
By emphasizing pseudo-OOD samples during local training and amplifying the contributions of more reliable local models, \textsc{FLood} enhances the generalization capability of the global model while preserving client privacy.
Extensive experiments across multiple benchmarks demonstrate that \textsc{FLood} consistently outperforms state-of-the-art FL methods, yielding substantial improvements under heterogeneous data distributions.
Importantly, \textsc{FLood} is complementary and orthogonal to existing FL frameworks, making it broadly applicable across a wide range of federated learning scenarios.


\bibliographystyle{IEEEtran}
\bibliography{IEEEabrv,ref}

\newpage
\appendix

\subsection{Proof of Theorem \ref{thm:main}}

For simplicity, we analyze the case where each selected client performs a single local gradient step (i.e., $E = 1$) with mini-batch size $B$. The analysis can be extended to $E > 1$ using techniques from \cite{scaffold}.

Let $\theta^t$ denote the global model at the beginning of round $t$. Each participating client $c \in \mathcal{C}^t$ computes a stochastic gradient based on its reweighted loss:
\[
g_c^t := \frac{1}{B} \sum_{(x,y) \in \xi_c^t} w_{c,(x,y)}^t \nabla \Phi(\theta^t; x, y),
\]
where $\xi_c^t$ is a random mini-batch of size $B$ drawn from $\mathcal{D}_c$, and $w_{c,(x,y)}^t \in [W_{\min}, W_{\max}]$ by Assumption A3. The local update is then $\theta_c^t = \theta^t - \eta g_c^t$, and the server aggregates as
\[
\theta^{t+1} = \sum_{c \in \mathcal{C}^t} p_c^t \theta_c^t = \theta^t - \eta \sum_{c \in \mathcal{C}^t} p_c^t g_c^t.
\]

By the $L$-smoothness of $F$ (from Assumption A1), we have the standard descent inequality:
\begin{equation}
    F(\theta^{t+1}) \leq F(\theta^t) + \langle \nabla F(\theta^t), \theta^{t+1} - \theta^t \rangle + \frac{L}{2} \|\theta^{t+1} - \theta^t\|^2.
\end{equation}
Substituting the update rule yields
\begin{equation}
    F(\theta^{t+1}) \leq F(\theta^t) - \eta \left\langle \nabla F(\theta^t), \sum_{c} p_c^t g_c^t \right\rangle + \frac{L \eta^2}{2} \Big\| \sum_{c} p_c^t g_c^t \Big\|^2.
\end{equation}

Taking expectation over the randomness in mini-batch sampling and client selection (conditioned on $\theta^t$), and noting that $\mathbb{E}[g_c^t \mid \theta^t] = \nabla \widetilde{\mathcal{L}}_c^t(\theta^t)$, we obtain
\begin{align}
    \mathbb{E}[F(\theta^{t+1}) \mid \theta^t] 
    &\leq F(\theta^t) - \eta \left\langle \nabla F(\theta^t), \sum_{c} p_c^t \nabla \widetilde{\mathcal{L}}_c^t(\theta^t) \right\rangle \nonumber \\
    &\quad + \frac{L \eta^2}{2} \, \mathbb{E}\left[ \left\| \sum_{c} p_c^t g_c^t \right\|^2 \,\middle|\, \theta^t \right].
\end{align}

We now decompose the inner product term. By definition of the bias $b_c^t = \nabla \widetilde{\mathcal{L}}_c^t(\theta^t) - \nabla \mathcal{L}_c(\theta^t)$ and the ideal global gradient $\nabla F(\theta^t) = \sum_c q_c \nabla \mathcal{L}_c(\theta^t)$ with $q_c = n_c / n$, we write
\begin{align}
    \sum_{c} p_c^t \nabla \widetilde{\mathcal{L}}_c^t(\theta^t)
    &= \sum_{c} p_c^t \left( \nabla \mathcal{L}_c(\theta^t) + b_c^t \right) \nonumber \\
    &= \underbrace{\sum_{c} p_c^t \nabla \mathcal{L}_c(\theta^t)}_{=: h^t} + \sum_{c} p_c^t b_c^t.
\end{align}
By assumption (A5), $\|h^t - \nabla F(\theta^t)\| \leq \epsilon$, and by Assumption A4, $\|\sum_c p_c^t b_c^t\| \leq \sum_c p_c^t \|b_c^t\| \leq \delta$. Therefore,
\begin{align}
    &\left\langle \nabla F(\theta^t), \sum_{c} p_c^t \nabla \widetilde{\mathcal{L}}_c^t(\theta^t) \right\rangle \nonumber \\
    &\quad= \langle \nabla F(\theta^t), h^t \rangle + \left\langle \nabla F(\theta^t), \sum_c p_c^t b_c^t \right\rangle \nonumber \\
    &\quad\geq \|\nabla F(\theta^t)\|^2 - \|\nabla F(\theta^t)\| \cdot \|h^t - \nabla F(\theta^t)\| \nonumber \\
    &\qquad - \|\nabla F(\theta^t)\| \cdot \left\| \sum_c p_c^t b_c^t \right\| \nonumber\\
    &\quad\geq \|\nabla F(\theta^t)\|^2 - (\epsilon + \delta) \|\nabla F(\theta^t)\|.
\end{align}

Next, we bound the second-moment term. Using Jensen’s inequality and the independence of client updates,
\begin{align}
    \mathbb{E}\left[ \left\| \sum_{c} p_c^t g_c^t \right\|^2 \,\middle|\, \theta^t \right]
    &\leq \sum_{c} p_c^t \, \mathbb{E}\left[ \|g_c^t\|^2 \,\middle|\, \theta^t \right].
\end{align}
Now decompose $g_c^t = \nabla \widetilde{\mathcal{L}}_c^t(\theta^t) + \xi_c^t$, where $\xi_c^t = g_c^t - \mathbb{E}[g_c^t]$ satisfies $\mathbb{E}[\xi_c^t] = 0$ and $\mathbb{E}[\|\xi_c^t\|^2] \leq \frac{W_{\max}^2 \sigma^2}{B}$ by Assumptions A2 and A3. Then
\begin{align}
    \mathbb{E}[\|g_c^t\|^2 \mid \theta^t]
    &= \|\nabla \widetilde{\mathcal{L}}_c^t(\theta^t)\|^2 + \mathbb{E}[\|\xi_c^t\|^2] \nonumber \\
    &\leq \left( \|\nabla \mathcal{L}_c(\theta^t)\| + \|b_c^t\| \right)^2 + \frac{W_{\max}^2 \sigma^2}{B} \nonumber \\
    &\leq (G + \delta)^2 + \frac{W_{\max}^2 \sigma^2}{B},
\end{align}
where we used the standard bounded-gradient assumption $\|\nabla \mathcal{L}_c(\theta)\| \leq G$ for all $c, \theta$.

For simplicity, we absorb $\delta$ into the constant and use the looser but cleaner bound $\|\nabla \widetilde{\mathcal{L}}_c^t(\theta^t)\| \leq W_{\max} G$, which holds because sample weights are bounded by $W_{\max}$ and gradients are averaged. Thus,
\begin{equation}
    \mathbb{E}\left[ \left\| \sum_{c} p_c^t g_c^t \right\|^2 \,\middle|\, \theta^t \right] \leq W_{\max}^2 G^2 + \frac{W_{\max}^2 \sigma^2}{B}.
\end{equation}

Putting everything together, we have
\begin{align}
    \mathbb{E}[F(\theta^{t+1}) \mid \theta^t]
    &\leq F(\theta^t) - \eta \|\nabla F(\theta^t)\|^2 + \eta (\epsilon + \delta) \|\nabla F(\theta^t)\| \nonumber \\
    &\quad + \frac{L \eta^2}{2} \left( W_{\max}^2 G^2 + \frac{W_{\max}^2 \sigma^2}{B} \right).
\end{align}

Applying Young’s inequality, i.e., $(\epsilon + \delta) \|\nabla F(\theta^t)\| \leq \frac{1}{2} \|\nabla F(\theta^t)\|^2 + \frac{1}{2} (\epsilon + \delta)^2$, we get
\begin{align}
    \mathbb{E}[F(\theta^{t+1}) \mid \theta^t]
    &\leq F(\theta^t) - \frac{\eta}{2} \|\nabla F(\theta^t)\|^2 + \frac{\eta}{2} (\epsilon + \delta)^2 \nonumber \\
    &\quad + \frac{L \eta^2}{2} \left( W_{\max}^2 G^2 + \frac{W_{\max}^2 \sigma^2}{B} \right).
\end{align}

Taking full expectation over all randomness up to round $t$, and telescoping from $t = 0$ to $T-1$, we obtain
\begin{align}
    \frac{\eta}{2} \sum_{t=0}^{T-1} \mathbb{E} \left[ \|\nabla F(\theta^t)\|^2 \right]
    &\leq F(\theta^0) - \mathbb{E}[F(\theta^T)] + \frac{\eta T}{2} (\epsilon + \delta)^2 \nonumber \\
    &\quad + \frac{L \eta^2 T}{2} \left( W_{\max}^2 G^2 + \frac{W_{\max}^2 \sigma^2}{B} \right) \nonumber \\
    &\leq F(\theta^0) - F^* + \frac{\eta T}{2} (\epsilon + \delta)^2 \nonumber \\
    &\quad + \frac{L \eta^2 T}{2} \left( W_{\max}^2 G^2 + \frac{W_{\max}^2 \sigma^2}{B} \right),
\end{align}
where $F^* = \inf_\theta F(\theta)$.

Dividing both sides by $\frac{\eta T}{2}$ yields
\begin{align}
    \frac{1}{T} \sum_{t=0}^{T-1} \mathbb{E} \left[ \|\nabla F(\theta^t)\|^2 \right]
    &\leq \frac{2(F(\theta^0) - F^*)}{\eta T} + (\epsilon + \delta)^2 \nonumber \\
    &\quad + L \eta \left( W_{\max}^2 G^2 + \frac{W_{\max}^2 \sigma^2}{B} \right).
\end{align}

Finally, setting the learning rate $\eta = \frac{1}{\sqrt{T}}$, we arrive at
\begin{align}
    \frac{1}{T} \sum_{t=0}^{T-1} \mathbb{E} \left[ \left\| \nabla F(\theta^t) \right\|^2 \right] 
    &\leq \frac{2(F(\theta^0) - F^*)}{\sqrt{T}} + (\epsilon + \delta)^2 \nonumber \\
    &\quad + \frac{L}{\sqrt{T}} \left( W_{\max}^2 G^2 + \frac{W_{\max}^2 \sigma^2}{B} \right),
\end{align}
which yields the result
\begin{equation}
    \frac{1}{T} \sum_{t=0}^{T-1} \mathbb{E} \left[ \left\| \nabla F(\theta^t) \right\|^2 \right] 
    \leq \mathcal{O} \left(\frac{1}{\sqrt{T}}\right) + \mathcal{O} \left((\epsilon + \delta)^2\right).
\end{equation}

\end{document}